\documentclass{ecai} 

\usepackage{latexsym}
\usepackage{amssymb}
\usepackage{amsmath}
\usepackage{amsthm}
\usepackage{booktabs}
\usepackage{enumitem}
\usepackage{graphicx}
\usepackage{color}

\usepackage{mathrsfs}
\usepackage{subcaption}
\usepackage{multirow}
\usepackage{url}

\usepackage{algorithm}
\usepackage{algpseudocode}

\usepackage{lipsum}
\usepackage{pdfpages}

\newtheorem{theorem}{Theorem}

\newtheorem{proposition}[theorem]{Proposition}

\def\eg{\emph{e.g. }} 

\def\ie{\emph{i.e. }}

\newcommand{\BibTeX}{B\kern-.05em{\sc i\kern-.025em b}\kern-.08em\TeX}

\begin{document}


\begin{frontmatter}


\paperid{4188} 


\title{Incorporating Pre-trained Diffusion Models in \\ Solving the Schrödinger Bridge Problem}



\author[A]{\fnms{Zhicong}~\snm{Tang}}
\author[B]{\fnms{Tiankai}~\snm{Hang}}
\author[C]{\fnms{Shuyang}~\snm{Gu}}
\author[D]{\fnms{Dong}~\snm{Chen}}
\author[A,D]{\fnms{Baining}~\snm{Guo}}


\address[A]{Tsinghua University}
\address[B]{Southeast University}
\address[C]{University of Science and Technology of China}
\address[D]{Microsoft Research Asia}


\begin{abstract}
This paper aims to unify Score-based Generative Models (SGMs), also known as Diffusion models, and the Schrödinger Bridge (SB) problem through three reparameterization techniques: Iterative Proportional Mean-Matching (IPMM), Iterative Proportional Terminus-Matching (IPTM), and Iterative Proportional Flow-Matching (IPFM). These techniques significantly accelerate and stabilize the training of SB-based models.
Furthermore, the paper introduces novel initialization strategies that use pre-trained SGMs to effectively train SB-based models. By using SGMs as initialization, we leverage the advantages of both SB-based models and SGMs, ensuring efficient training of SB-based models and further improving the performance of SGMs.
Extensive experiments demonstrate the significant effectiveness and improvements of the proposed methods. We believe this work contributes to and paves the way for future research on generative models.
\end{abstract}

\end{frontmatter}

\section{Introduction}
\label{sec:intro}

Score-based Generative Models (SGMs)~\cite{song2019generative,ho2020denoising,lipman2022flow}, also known as Diffusion models, have recently achieved remarkable success~\cite{dhariwal2021diffusion,rombach2022high,gu2022vector}. Despite their advancements, SGMs require carefully crafted noise schedules and tailored to diverse tasks~\cite{karras2022elucidating,lin2024common,chen2023importance}. Without customizations, SGMs may encounter significant difficulties in handling complex data, including video~\cite{blattmann2023stable,ho2022imagen,sora2024} and 3D content~\cite{jun2023shap,tang2023volumediffusion}. Moreover, SGMs are limited to pre-defined distributions, \eg Gaussian distributions~\cite{ho2020denoising}, which narrows further applications in a wider range of scenarios, such as conditional generation tasks like unpaired domain transfer~\cite{zhu2017unpaired}.

The Schrödinger Bridge (SB)~\cite{schrodinger1932theorie,follmer1988random} arises as a more generalized framework that constructs the transition between two arbitrary distributions. Numerous implementations of SB have been proposed~\cite{shi2024diffusion,bernton2019schr,chen2016entropic,caluya2021wasserstein,pavon2021data}, and one theoretical solution is the Iterative Proportional Fitting (IPF)~\cite{fortet1940resolution,kullback1968probability}. Nevertheless, practical application remains challenging because IPF involves optimizing the joint distribution of forward and backward trajectories, which can be highly complex. 

Diffusion Schrödinger Bridge (DSB)~\cite{de2021diffusion} further simplifies and approximates IPF by optimizing a series of conditional distributions instead of the whole joint distribution. DSB employs two neural networks to symmetrically transit from one distribution to the other with iterative training. Iterative Proportional Maximum Likelihood (IPML)~\cite{vargas2021solving} develops solutions originated from maximum-likelihood-based method. Recent work, Diffusion Schrödinger Bridge Matching (DSBM)~\cite{shi2024diffusion}, also solves SB by iteratively using Markovian and Reciprocal Projection.

\begin{figure}[t]
    \centering
    \includegraphics[width=0.9\columnwidth]{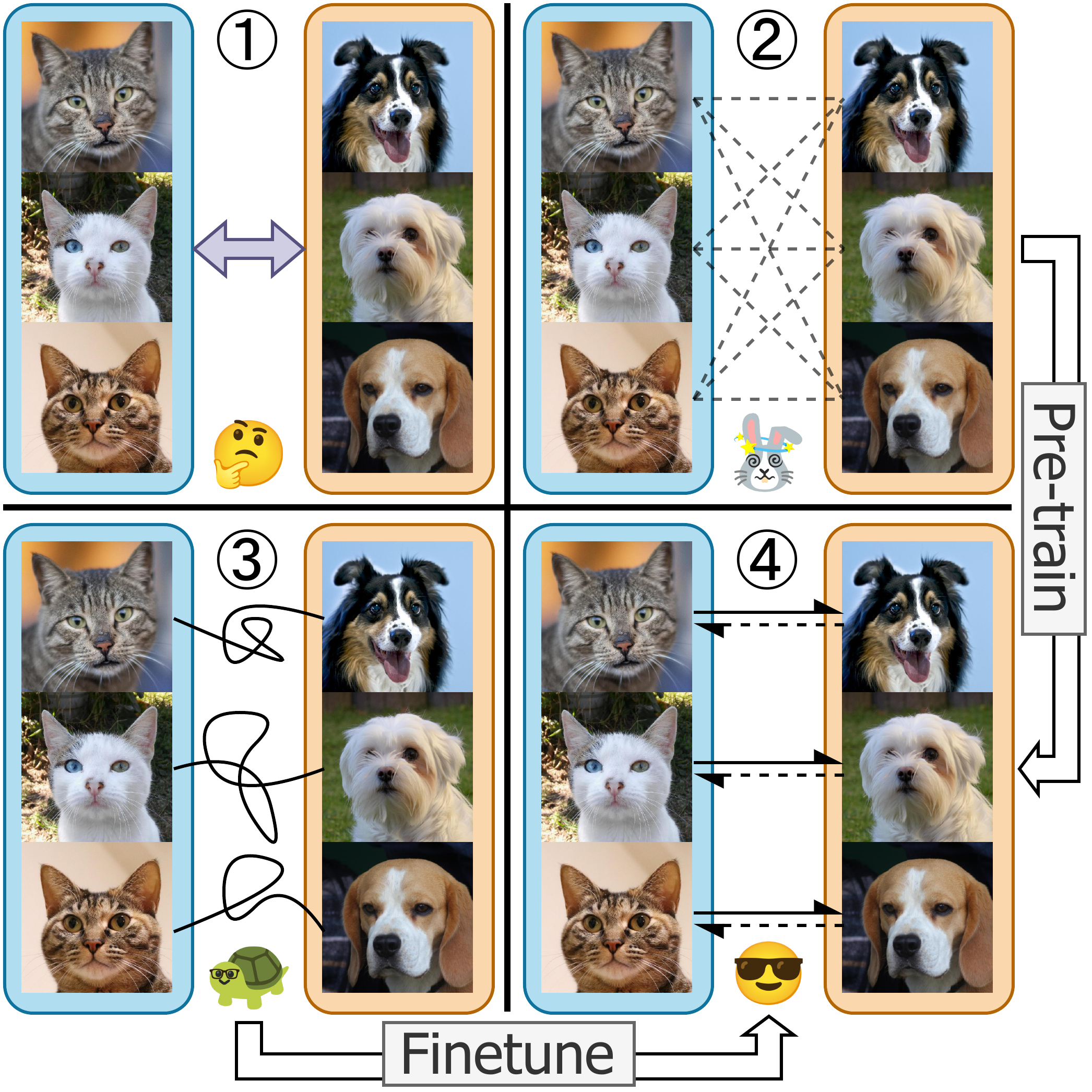}
    \vskip 0.25cm
    \caption{(1) Schrödinger Bridge (SB) aims to automatically solve the connections between two distributions without given pairing rules. (2) Diffusion models struggle to determine the correct pairings, as the training pairs are randomly selected. (3) SB-based methods dynamically learn the connections, but are extremely slow and unstable. (4) By using diffusion models as pre-train initialization and further finetuning with SB, our method solve the connections with greater speed, stability, and quality.}
    \label{fig:teasor}
    \vspace{0.5cm}
\end{figure}

While these methods~\cite{de2021diffusion,vargas2021solving,shi2024diffusion} have made great progress, SB-based models still struggle to scale up to larger scales, such as high-resolution and natural images. They suffer from long and unstable training processes and consume extensive computation. On the other hand, SGMs have been widely utilized and numerous off-the-shelf pre-trained models with superior quality are available, such as Stable Diffusion~\cite{rombach2022high}. Various advanced designs and techniques, such as noise schedule~\cite{chen2023importance}, Ordinary Differential Equation (ODE) based interpolants~\cite{lipman2022flow}, and bespoke solvers~\cite{lu2022dpm}, have greatly boosted the capability of SGMs. Research on leveraging beneficial designs from SGMs and connecting the field of SB and SGMs still remains empty.

As shown in Figure~\ref{fig:teasor}, this work aims to bridge the gap between SB and SGMs, and proposes novel methods for incorporating pre-trained SGMs in solving the SB problem. We propose three reparameterization techniques that unify the training of SB-based models and SGMs, including \emph{Iterative Proportional Mean-Matching (IPMM)}, \emph{Iterative Proportional Terminus-Matching (IPTM)}, and \emph{Iterative Proportional Flow-Matching (IPFM)}. Moreover, our theoretical analysis reveals that initialization is pivotal for solving SB, addressing the issue of slow convergence and yielding improved performance. Therefore, we develop initialization methods that use powerful pre-trained SGMs to greatly accelerate and stabilize the training of SB-based models, based on these reparameterization techniques.

We conduct comprehensive experiments to confirm the effectiveness and the generation quality of the proposed methods. Our contributions are summarized as below:

\begin{itemize}
    \item[•] We introduce three reparameterization techniques for SB-based models, including IPMM, IPTM, and IPFM, to bridge the gap between SB and SGMs.
    \item[•] We propose initialization strategies that successfully integrate pre-trained SGMs into training SB-based models.
    \item[•] We demonstrate the effectiveness of our methods through extensive experiments on synthetic datasets, image generation, and unpaired image translation.
\end{itemize}

\section{Preliminary}

\noindent \textbf{Notation} Consider a data distribution $p_\textup{data}$ and a prior distribution $p_\textup{prior}$ in $\mathbb{R}^d$. Denote the transition from $p_\textup{data}$ to $p_\textup{prior}$ as the forward process, and from $p_\textup{prior}$ to $p_\textup{data}$ as the backward process. In discrete time horizons $k\in\{0,\dots,N\}$, $\mathscr{P}_k=\mathscr{P}\left((\mathbb{R}^d)^k\right)$ denotes the space of $k$-state joint distributions for any $k\in\mathbb{N}_{\le N+1}$. In continuous setting $t\in[0,T]$, we use $\mathscr{P}(\mathcal{C})$ where $\mathcal{C}=\textup{C}\left([0,T],\mathbb{R}^d\right)$. 

\subsection{Score-based Generative Models}
\label{sec:SGM}

Score-based Generative Models (SGMs)~\cite{song2019generative,ho2020denoising}, also known as Diffusion models, connect two distributions through a dual process that can be modeled as Markov chains. Given $p_{\textup{data}}$ and $p_{\textup{prior}}$, the forward process $p_{k+1|k}(x_{k+1}|x_k)$ transits $p_0 = p_{\textup{data}}$ to $p_N\approx p_{\textup{prior}}$. The forward process generates a sequence $x_{0:N}$ with the joint density
\begin{equation}
    p(x_{0:N})=p_0(x_0)\prod_{k=0}^{N-1}p_{k+1|k}(x_{k+1}|x_k).
    \label{eq:forward_chain}
\end{equation}

\noindent The joint density can also be formulated as a time reversal:
\begin{equation}
    \begin{aligned}
        & p(x_{0:N})=p_N(x_N)\prod_{k=0}^{N-1}p_{k|k+1}(x_k|x_{k+1}),\\
        &\text{where}\ p_{k|k+1}(x_k|x_{k+1})=\frac{p_{k+1|k}(x_{k+1}|x_k)p_k(x_k)}{p_{k+1}(x_{k+1})}.
    \end{aligned}
    \label{eq:backward_chain}
\end{equation}

However, directly computing $p_{k|k+1}(x_k|x_{k+1})$ is typically challenging. SGM utilizes a simplified approach that regards the forward process as gradually adding Gaussian noise
\begin{equation}
    p_{k+1|k}(x_{k+1}|x_k)=\mathcal{N}(x_{k+1};x_k+\gamma_{k+1}f_k(x_k),2\gamma_{k+1}\mathbf{I}),
    \label{eq:forward_gaussian}
\end{equation}

\noindent where $f_k(x_k)$ is the drift term and $\gamma_k$ determines the noise strengths of each step. It follows that for a sufficiently large $N$, the distribution $p_N$ will converge to Gaussian, denoted as $p_{\textup{prior}}$. Moreover, the backward transition in Equation~(\ref{eq:backward_chain}) can be analytically approximated~\cite{vincent2011connection,anderson1982reverse} as
\begin{equation}
    \begin{aligned}
        &\ p_{k|k+1}(x_k|x_{k+1})\\
        =&\ p_{k+1|k}(x_{k+1}|x_k)\exp\left(\log p_k(x_k)-\log p_{k+1}(x_{k+1})\right)\\
        \approx&\ \mathcal{N}(x_k;x_{k+1}-\gamma_{k+1}f_{k+1}(x_{k+1})+ \\
        &\quad\quad\quad 2\gamma_{k+1}\nabla\log p_{k+1}(x_{k+1}),2\gamma_{k+1}\mathbf{I}).
    \end{aligned}
    \label{eq:backward_gaussian}
\end{equation}

\noindent using that $p_k\approx p_{k+1}$, a Taylor expansion of $\log p_{k+1}$ at $x_{k+1}$ and $f(x_k)\approx f(x_{k+1})$. Subsequently, SGM employs neural networks $s_\theta(x_{k+1},k+1)$ to approximate the score term $\nabla\log p_{k+1}(x_{k+1})$. Then, we can sample $x_0 \sim p_{\textup{data}}$ by sampling $x_N \sim p_{\textup{prior}}$ and iterate $x_k \sim p_{k|k+1}(x_k|x_{k+1})$.

\subsection{Schrödinger Bridge and Iterative Proportional Fitting}
\label{sec:SB_and_DSB}

Consider a reference density $p_\textup{ref} \in \mathscr{P}_{N+1}$ given by Equation~(\ref{eq:forward_chain}), the Schrödinger Bridge (SB) problem aims to find $\pi^* \in \mathscr{P}_{N+1}$ which satisfies
{\small
\begin{equation}
    \pi^*=\arg\min\{\textup{KL}(\pi|p_\textup{ref}):\pi\in\mathscr{P}_{N+1},\pi_0=p_\textup{data},\pi_N=p_\textup{prior}\}.
    \label{eq:SB_definition}
\end{equation}
}

Generally, the SB problem lacks a closed-form solution. Researchers employ the Iterative Proportional Fitting (IPF)~\cite{fortet1940resolution,kullback1968probability} to address it through iterative optimization:
\begin{equation}
    \label{eq:IPF_definition}
    \begin{aligned}
        \pi^{2n+1}&=\arg\min\{\textup{KL}(\pi|\pi^{2n}):\pi\in\mathscr{P}_{N+1},\pi_N=p_\textup{prior}\},\\
        \pi^{2n+2}&=\arg\min\{\textup{KL}(\pi|\pi^{2n+1}):\pi\in\mathscr{P}_{N+1},\pi_0=p_\textup{data}\}.
    \end{aligned}
\end{equation}

\noindent where $\pi^0=p_\textup{ref}$ is the initialization condition. Nevertheless, IPF needs to compute and optimize the joint density, which is usually infeasible within practical settings due to computational complexity.

\section{Related works}

\subsection{Image-to-image translation}

The first image-to-image translation framework, Pix2Pix~\cite{isola2017image}, is based on conditional GANs. Later, CycleGAN~\cite{zhu2017unpaired} and DualGAN~\cite{yi2017dualgan} separately used two GANs on two domains and trained them together with dual learning~\cite{he2016dual}, thereby enabling learning from unpaired data. Other GAN-based techniques for image-to-image translation, such as unsupervised cross-domain~\cite{taigman2016unsupervised}, multi-domain~\cite{choi2020stargan}, and few-shot~\cite{liu2019few} methods have also been suggested. However, GAN-based methods severely suffer from instability and mode collapse, and can not generalize to complex distributions.

Alternatively, other works attempt to approximate the boundary distribution as a mixture of Dirac deltas. I$^2$SB~\cite{liu20232} was the first to adopt this in image-to-image translations. Bridge-TTS~\cite{chen2023schrodinger} studies the tractable SB between Gaussian-smoothed paired data. BBDM~\cite{li2023bbdm} utilizes diffusion models to fit the Brownian Bridge process between training pairs. DDBM~\cite{zhou2023denoising} also transforms diffusion models by injecting Stochastic bridges with $h$-transform. However, these methods can only be applied to paired datasets, rely on explicitly given pairs from training sets, and do not generalize to unpaired tasks.

\subsection{Diffusion Schrödinger Bridge}
\label{sec:related_DSB}

Diffusion Schrödinger Bridge (DSB)~\cite{de2021diffusion} can be conceptualized as an approximation of IPF. It dissects the joint density $\pi$ into optimizing conditional densities $\pi_{k+1|k}$ and $\pi_{k|k+1}$ with
{\small
\begin{equation}
    \label{eq:DSB_definition}
    \begin{aligned}
        \pi^{2n+1}&=\arg\min\{\textup{KL}(\pi_{k|k+1}|\pi^{2n}_{k|k+1}):\pi\in\mathscr{P}_{N+1},\pi_N=p_\textup{prior}\},\\
        \pi^{2n+2}&=\arg\min\{\textup{KL}(\pi_{k+1|k}|\pi^{2n+1}_{k+1|k}):\pi\in\mathscr{P}_{N+1},\pi_0=p_\textup{data}\}.
    \end{aligned}
\end{equation}
}
 
In addition, DSB follows SGMs to assume the conditional transitions $\pi_{k+1|k}$ and $\pi_{k|k+1}$ to be Gaussian, in order to analytically calculate the time reversal as Equation~(\ref{eq:backward_gaussian}). Then DSB employs two separate neural networks to model $\pi_{k+1|k}$ and $\pi_{k|k+1}$ with the training objective
{\small
\begin{equation}
    \label{eq:DSB_original_loss}
    \begin{aligned}
        &\mathcal{L}_{B_{k+1}^n}=\mathbb{E}_{ p_{k,k+1}^n}\left[\left\|B_{k+1}^n(x_{k+1})-\left(x_{k+1}+F_k^n(x_k)-F_k^n(x_{k+1})\right)\right\|^2\right],\\
        &\mathcal{L}_{F_k^{n+1}}=\mathbb{E}_{q_{k,k+1}^n}\left[\left\|F_{k}^{n+1}(x_{k})-\left(x_{k}+B_{k+1}^n(x_{k+1})-B_{k+1}^n(x_{k})\right)\right\|^2\right],
    \end{aligned}
\end{equation}
}

\noindent denoting $p^n=\pi^{2n}$ and $q^n=\pi^{2n+1}$. $p_{k+1|k}^n(x_{k+1}|x_{k}) = \mathcal{N}(x_{k+1};x_k+\gamma_{k+1}f_{k}^n(x_{k}), 2\gamma_{k+1} I)$ is the forward transition from $p_\textup{data}$ to $p_\textup{prior}$ in accordance to Equation~(\ref{eq:forward_gaussian}). On the other side, the backward transition is $q_{k|k+1}^n(x_k|x_{k+1}) = \mathcal{N}(x_k;x_{k+1}+\gamma_{k+1}b_{k+1}^n(x_{k+1}), 2\gamma_{k+1} I)$ from $p_\textup{prior}$ to $p_\textup{data}$.

In practice, DSB approximates $B_{\beta^n}(k,x)\approx B_{k}^n(x)=x+\gamma_kb_k^n(x)$ and $F_{\alpha^n}(k,x)\approx F_{k}^n(x)=x+\gamma_{k+1}f_{k}^n(x)$ with two neural networks $\alpha$ and $\beta$, respectively. With $p_0=p_\textup{ref}$ manually pre-defined in Equation~(\ref{eq:forward_gaussian}), it iteratively optimizes $B_{\beta^n}$ and $F_{\alpha^n}$ subject to Equation~(\ref{eq:DSB_original_loss}). We denote $\bar{\gamma}_k=\sum_{i=0}^{k}\gamma_k$ and refer to the optimization of $B_{\beta^n}$ and $F_{\alpha^n}$ as the $(2n+1)^\textup{th}$ epoch and the $(2n)^\textup{th}$ epoch, respectively.

\subsection{Other Schrödinger Bridge based methods}

Recent methods have emerged to solve the SB problem by discretizing the state-space~\cite{chen2016entropic}, approximating potential functions with regression~\cite{bernton2019schr}, performing kernel density estimation~\cite{pavon2021data}, or using a Gaussian mixture model~\cite{korotin2023light}. SB-FBSDE~\cite{chen2021likelihood} proposes maximizing likelihood using forward-backward SDEs theory. DSBM~\cite{shi2024diffusion} solves the problem by iteratively applying Markovian and Reciprocal Projection. IPML~\cite{vargas2021solving} proves that solving the SB problem is equivalent to an autoregressive maximum likelihood estimation objective, which shares a similar form with our IPMM. However, these studies did not discuss or leverage pre-trained SGMs.

\section{Reparameterizing network prediction}
\label{sec:simplified_target}

To incorporate pre-trained SGMs as initialization, we need to reformulate the training objective to adapt to the form of prevalent SGMs. We introduce three types of reparameterization techniques for different classes of SGMs. We also provide the proof of the theoretical convergence and equivalence of these training objectives.

\subsection{Mean-matching objective}

Unlike DSB~\cite{de2021diffusion}, we opt to use an alternative objective as
\begin{equation}
    \begin{aligned}
        &\mathcal{L}_{B_{k+1}^n}^{'}=\mathbb{E}_{ p_{k,k+1}^n}\left[\left\|B_{k+1}^n(x_{k+1})-x_{k}\right\|^2\right], \\
        &\mathcal{L}_{F_k^{n+1}}^{'}=\mathbb{E}_{ q_{k,k+1}^n}\left[\left\|F_{k}^{n+1}(x_{k})-x_{k+1}\right\|^2\right],
    \end{aligned}
    \label{eq:DSB_simplified_loss}
\end{equation}

\noindent where networks learn to predict the mean of the next states on Markov chains. We name it as \emph{Iterative Proportional Mean-Matching (IPMM)}. It plays an important role in connecting SGMs and SB-based models, which is detailed in Section~\ref{sec:incorporating_pretrained_models}. It halves the number of forward evaluations (NFE) comparing to the original form of DSB in Equation~(\ref{eq:DSB_original_loss}), and greatly accelerates the training process.

\subsection{Terminus-matching objective}
\label{sec:terminus_matching}

Moreover, we draw inspirations from prevalent SGMs and introduce additional reparameterized objectives to further stabilize and accelerate the training of SB-based models. Following DDPM~\cite{ho2020denoising}, we propose \emph{Iterative Proportional Terminus-Matching (IPTM)}, which uses neural networks to predict the terminus of sampling trajectories. The training objective is 
\begin{equation}
    \begin{aligned}
        &\mathcal{L}_{\tilde{B}_{k+1}^n}=\mathbb{E}_{p_{0,k+1}^n}\left[\left\|\tilde{B}_{k+1}^n(x_{k+1})-x_{0}\right\|^2\right], \\
        &\mathcal{L}_{\tilde{F}_{k}^{n+1}}=\mathbb{E}_{q_{k,N}^n}\left[\left\|\tilde{F}_{k}^{n+1}(x_{k})-x_{N}\right\|^2\right].
    \end{aligned}
    \label{eq:TRDSB_loss}
\end{equation}

\noindent IPTM connects to IPMM via the transformation 
\begin{equation}
    \begin{aligned}
        &B_{k+1}^n(x_{k+1})=x_{k+1}+\frac{\gamma_{k+1}}{\bar{\gamma}_{k+1}}(\tilde{B}_{k+1}^n(x_{k+1})-x_{k+1}), \\
        &F_{k}^{n+1}(x_{k})=x_k+\frac{\gamma_{k+1}}{1-\bar{\gamma}_{k}}(\tilde{F}_{k}^{n+1}(x_{k})-x_k).
    \end{aligned}
    \label{eq:TRDSB_loss_1}
\end{equation}

\subsection{Flow-matching objective}
\label{sec:flow_matching}

Similar to Flow Matching~\cite{lipman2022flow}, we propose \emph{Iterative Proportional Flow-Matching (IPFM)}, which employs neural networks to predict the vector that connects the current state to the end state. The training objective is
\begin{equation}
    \begin{aligned}
        &\mathcal{L}_{\tilde{b}_{k+1}^n}=\mathbb{E}_{p_{0,k+1}^n}\left[\left\|\tilde{b}_{k+1}^n(x_{k+1})-\frac{x_0-x_{k+1}}{\bar{\gamma}_{k+1}}\right\|^2\right], \\
        &\mathcal{L}_{\tilde{f}_{k}^{n+1}}=\mathbb{E}_{q_{k,N}^n}\left[\left\|\tilde{f}_{k}^{n+1}(x_{k})-\frac{x_N-x_{k}}{1-\bar{\gamma}_{k}}\right\|^2\right].
    \end{aligned}
    \label{eq:FRDSB_loss}
\end{equation}

\noindent IPFM also connects to IPMM via the transformation 
\begin{equation}
    \begin{aligned}
        &B_{k+1}^n(x_{k+1})=x_{k+1}+\gamma_{k+1}\tilde{b}_{k+1}^n(x_{k+1}), \\
        &F_{k}^{n+1}(x_{k})=x_k+\gamma_{k+1}\tilde{f}_{k}^{n+1}(x_{k}).
    \end{aligned}
    \label{eq:FRDSB_loss_1}
\end{equation}

\subsection{Convergence and equivalence}

These reparameterized training objectives not only bridges the gap between SB and SGMs, accelerates and stabilize the training of SB, but also possesses the theoretical guarantee of convergence through their equivalence to DSB and IPF. The following proposition demonstrates that the proposed IPMM, IPTM, and IPFM are equivalent to the original form in Equation~(\ref{eq:DSB_original_loss}):

\begin{proposition}
    \label{prop:SDSB}
    Assume that for any $n\in\mathbb{N}$ and $k\in\{0,1,\ldots,N-1\}$, $\gamma_{k+1}>0$, $q_{k|k+1}^n=\mathcal{N}\left(x_k;B_{k+1}^n(x_{k+1}),2\gamma_{k+1}\mathbf{I}\right)$, $p_{k+1|k}^n=\mathcal{N}\left(x_{k+1};F_{k}^n(x_{k}),2\gamma_{k+1}\mathbf{I}\right)$, where $B_{k+1}^n(x)=x+\gamma_{k+1}b_{k+1}^n(x)$ and $F_{k}^n(x)=x+\gamma_{k+1}f_{k}^n(x)$. Similar to Equation~(\ref{eq:backward_gaussian}), we have
    \begin{equation}
        \begin{aligned}
            \mathcal{L}_{B_{k+1}^n}^{'}\approx\mathcal{L}_{B_{k+1}^n}, 
            \mathcal{L}_{F_k^{n+1}}^{'}\approx\mathcal{L}_{F_k^{n+1}}.
        \end{aligned}
        \label{eq:propositon_loss_equivalent}
    \end{equation}
\end{proposition}

\begin{proposition}
    Assume $\sum_{k=1}^N\gamma_{k}=1$. Given $x_{0:N}\sim p^n(x_{0:N})=p_0^n(x_0)\prod_{k=0}^{N-1}p_{k+1|k}^n(x_{k+1}|x_k)$ as the forward trajectories and $x_{0:N}'\sim q^n(x_{0:N})=q_N^n(x_N)\prod_{k=0}^{N-1}p_{k|k+1}^n(x_k|x_{k+1})$ as the backward trajectories. Under mild assumptions, we have
    \begin{equation}
        \begin{aligned}
            q_{k|k+1}^n(x_{k}|x_{k+1})&\approx p_{k|k+1,0}^n(x_{k}|x_{k+1},x_0) \\
            &=\mathcal{N}(x_{k};\mu_{k+1}^n(x_{k+1},x_0),\sigma_{k+1}\mathbf{I}),\\
            p_{k+1|k}^{n+1}(x_{k+1}|x_{k})&\approx q_{k+1|k,N}^n(x_{k+1}'|x_k',x_N') \\
            &=\mathcal{N}(x_{k+1}';\tilde{\mu}_k^n(x_k',x_N'),\tilde{\sigma}_{k+1}\mathbf{I}),
        \end{aligned}
        \label{eq:RDSB_1}
    \end{equation}
    where
    \begin{equation}
        \begin{aligned}
            & \mu_{k+1}^n(x_{k+1},x_0)\approx x_{k+1}+\frac{\gamma_{k+1}}{\bar{\gamma}_{k+1}}(x_0-x_{k+1}), \sigma_{k+1}=\frac{2\gamma_{k+1}\bar{\gamma}_{k}}{\bar{\gamma}_{k+1}}, \\
            & \tilde{\mu}^n_k(x_k',x_N')\approx x_k'+\frac{\gamma_{k+1}}{1-\bar{\gamma}_{k}}(x_N'-x_k'), \tilde{\sigma}_{k+1}=\frac{2\gamma_{k+1}(1-\bar{\gamma}_{k+1})}{1-\bar{\gamma}_k}.
        \end{aligned}
        \label{eq:RDSB_2}
    \end{equation}
    \label{prop:RDSB_prop}
\end{proposition}

We leave the detailed proof in Appendix B and Appendix C. These reparemeterized objectives provide several advantages. First, they save half of the number of forward evaluations (NFE) when computing the prediction targets. With Equation~(\ref{eq:DSB_original_loss}), it needs to run model twice to to compute one prediction target. For example, it consumes two network forwards $F_k^n(x_k)$ and $F_k^n(x_{k+1})$ to train $B_{k+1}^n(x_{k+1})$ once. On the other hand, our proposed simplified objectives in Equation~(\ref{eq:DSB_simplified_loss}), Equation~(\ref{eq:TRDSB_loss}), and Equation~(\ref{eq:FRDSB_loss}) need only one evaluation per target during sampling trajectories and eliminate the unnecessary $F_k^n(x_{k+1})$. This is critical when the networks $B_{k+1}^n(x)$ and $F_k^n(x)$ are large or the dimension of $x$ is high in practical settings, \eg high-resolution natural image generation.

Second, they provide accurate prediction targets for training and thus significantly stabilize the training process. In Equation~(\ref{eq:DSB_simplified_loss}), the networks benefits from the clear update direction that transits from one intermediate state on trajectories to the next one. In contrast, Equation~(\ref{eq:DSB_original_loss}) involves additional offsets like $F_k^n(x_k)-F_k^n(x_{k+1})$ and $B_{k+1}^n(x_{k+1})-B_{k+1}^n(x_{k})$. Moreover, instead of fitting intermediate states, Equation~(\ref{eq:TRDSB_loss}) and Equation~(\ref{eq:FRDSB_loss}) offers to match the terminus of trajectories and the direction vector to the terminus. Note that the trajectories are reversely sampled from two data distributions, the terminus are always valid samples like real images. The training can be remarkably accelerated and stabilized by fitting data samples rather than the outputs of another network, which is also being alternatively trained.

Finally, and of greater significance, these reparemeterized objectives are aligned with SGM. Whether training forward or backward models, the targets are simply to predict the subsequent state, akin to the approach of SGMs. This unification is crucial for the effective training of SB and enables us to incorporate pre-trained SGMs.

\begin{figure}[!t]
    \centering
    \includegraphics[width=0.8\columnwidth]{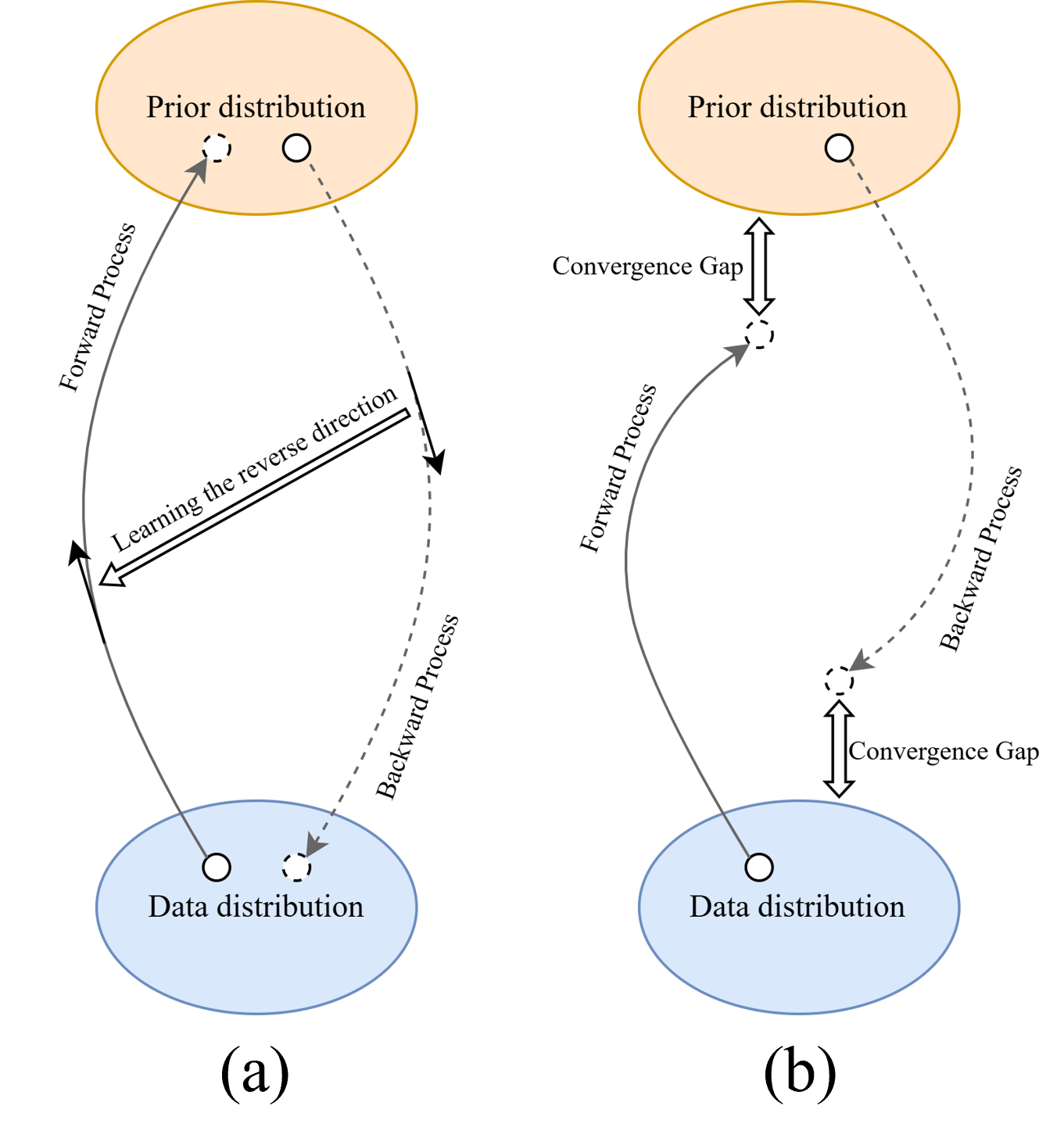}
    \caption{Illustration of the alternative training process. (a) Given the forward process that transits from data to prior (solid line), the backward model transits from prior back to data (dashed line) by learning the reverse direction on trajectory. (b) Since the training trajectories and targets are derived from previous models, the backward model struggles to converge if the forward process is poorly initialized.}
    \label{fig:sb_init}
    \vspace{0.7cm}
\end{figure}

\section{Incorporating pre-trained SGMs}
\label{sec:incorporating_pretrained_models}

In this section, we first analyze the convergence of DSB, and point out that the alternative training strategy is recursively relied on the convergence of previous round. This explains the instability of its training process and reveals that the proper initialization is the key for the training. Next, we propose novel methods to incorporate pre-trained SGMs as powerful initialization for SB in light of the proposed reparemeterized objectives. We also discuss that the initialization method is only applicable to our IPMM, IPTM, and IPFM.

\subsection{Convergence analysis}
\label{sec:convergence_analysis}

We illustrate the training framework of DSB in Figure~\ref{fig:sb_init}. While SB solves the transition between two arbitrary distributions, we use $p_\textup{data}$ and $p_\textup{prior}$ to refer to these two distributions for clarity and in order to be consistent with SGM. Specifically, for each epoch, the backward model is trained to map from $p_\textup{prior}$ back to $p_\textup{data}$, and the forward model is trained to approximate the mapping from $p_\textup{data}$ to $p_\textup{prior}$.

During practical training, we observed that the results of DSB usually exhibit slow convergence and requiring an excessive number of training iterations to reach optimal state. To elucidate the underlying cause, we conducted an analysis of the convergence performance and take the backward process as an example. Note that $p_\textup{data}=\pi_0^{2n}=p_0^n=\int p_N(x_N)\prod_{i=0}^{N-1}p_{i|i+1}^n(x_i|x_{i+1})\mathrm{d}x_i$, we have

\begin{equation}
    \begin{aligned}
        &\ \pi_0^{2n+1} - p_\textup{data} \\
        =&\ q_{0}^n(x_0) - p_\textup{data} \\
        =& \int q_{0|1}^n(x_0|x_1)q_{1}^n(x_1)\mathrm{d}x_1 - p_\textup{data} \\
        \vdots\ & \\
        =& \int q_{N}^n(x_N)\prod_{i=0}^{N-1}q_{i|i+1}^n(x_i|x_{i+1})\mathrm{d}x_i \\
        &\quad-\int p^n_N(x_N)\prod_{i=0}^{N-1}p_{i|i+1}^n(x_i|x_{i+1})\mathrm{d}x_i\\
        \stackrel{\text{*}}{=}& \int (p_\textup{prior}(x_N) - p^n_N(x_N)) \prod_{i=0}^{N-1}p_{i|i+1}^n(x_i|x_{i+1})\mathrm{d}x_i, \\
    \end{aligned}
    \label{eq:convergence_1}
\end{equation}

\noindent where $(*)$ is established that Equation~(\ref{eq:DSB_definition}) assumes every epoch achieves the completely converged state, \ie $q_{i|i+1}^n(x_i|x_{i+1})=p_{i|i+1}^n(x_i|x_{i+1})$ for every $i\in\{0,1,\ldots,N-1\}$. Similarly, we have

\begin{equation}
    \begin{aligned}
        \pi_N^{2n+2}-p_\textup{prior}=\int \left(q_\textup{data}(x_0)-q^n_0(x_0)\right)\prod_{i=1}^{N}q_{i+1|i}^n(x_{i+1}|x_{i})\mathrm{d}x_i.
    \end{aligned}
    \label{eq:convergence_2}
\end{equation}

Equation~(\ref{eq:convergence_1}) and Equation~(\ref{eq:convergence_2}) indicate that the convergence of the $(n+1)$-th epoch relies on the previous $n$-th epoch, underscoring the stability of training process. Models suffer from inaccurate trajectories with random or poor initialization, for example, Figure~\ref{fig:sb_init}(b), while training with proper initialization in Figure~\ref{fig:sb_init}(a) leads to rapid convergence. These precious insights suggest that an effective initial state is crucial in solving the SB problem.

\begin{algorithm}[!t]
    \caption{Training with IPFM and SGMs initialization}
    \hspace*{\algorithmicindent} \textbf{Input}: Pre-trained SGM $m_{\theta_1}(k,x)$, Training epochs $2L$, \\
    \hspace*{\algorithmicindent} \quad Timesteps $N$, Learning rate $\eta$, Data $p_\textup{data}$, Prior $p_\textup{prior}$ \\
    \hspace*{\algorithmicindent} \textbf{Output}: Forward network $F_{\alpha^{L}}$, Backward network $B_{\beta^{L}}$
    \begin{algorithmic}[1] 
      \For {$n\in\{1,\dots,L\}$}
        \If {$n=1$}
            \State Initialize with $B_{\beta^{0}}(k,x):=x+\frac{1}{N}m_{\theta_1}(k,x)$
        \EndIf
        \While {not converged}
            \State Sample trajectories $\{X^j_{k}\}_{k,j=0}^{N,M}$ for training, where $X^j_{k-1}=B_{\beta^{n-1}}(k, X^{j}_k)+\sqrt{2\gamma_{k}}\epsilon$ and $X^j_N \sim p_\textup{prior}$ 
            \State \textcolor{blue}{$\hat{\ell}_{n}^f\leftarrow\nabla_{\alpha^{n}}\left[\left\|F_{\alpha^{n}}(k,X^{j}_{k})-\frac{X^{j}_{N}-X^{j}_{k}}{1-\bar{\gamma}_{k}}\right\|^2\right]$}
            \State $\alpha^{n} \leftarrow \textrm{Gradient\_Step}(\hat{\ell}_{n}^f(\alpha^{n}))$
        \EndWhile
        \While {not converged}
            \State Sample trajectories $\{X^j_{k}\}_{k,j=0}^{N,M}$ for training, where $X^{j}_{k+1} = F_{\alpha^n}(k, X^{j}_{k})+\sqrt{2\gamma_{k+1}}\epsilon$ and $X^j_0 \sim p_\textup{data}$
            \State \textcolor{blue}{$\hat{\ell}_n^b\leftarrow\nabla_{\beta^n}\left[\left\|B_{\beta^n}(k+1,X^{j}_{k+1})-\frac{X^{j}_{0}-X^{j}_{k+1}}{\bar{\gamma}_{k+1}}\right\|^2\right]$}
            \State $\beta^{n} \leftarrow \textrm{Gradient\_Step}(\hat{\ell}_n^b(\beta^n))$ 
        \EndWhile
      \EndFor
    \end{algorithmic}
    \label{alg:SDSB}
\end{algorithm}

\subsection{SGMs as plug-and-play initialization}

Next, we show how to use pre-trained SGMs as powerful initialization for SB-based models. For simplicity, consider we have two pre-trained SGMs~\cite{lipman2022flow}, denoted as $\theta_1$ and $\theta_2$. $N$ denotes the number of total timesteps, and $k$ is index of an intermediate timestep. $\theta_1$ transits from $p_\textup{prior}$ to $p_\textup{data}$, \ie the model $m_{\theta_1}$ takes $x_k=\left(1-\frac{k}{N}\right)x_0+\frac{k}{N}x_1$, where $x_0\sim p_\textup{data}$ and $x_1\sim p_\textup{prior}$, and learns $m_{\theta_1}(k,x_k)\rightarrow x_0-x_1$. On the other hand, $\theta_2$ transits from $p_\textup{data}$ to $p_\textup{prior}$, \ie the model $m_{\theta_2}$ takes $x_k=\frac{k}{N}x_0+\left(1-\frac{k}{N}\right)x_1$ and learns $m_{\theta_2}(k,x_k)\rightarrow x_1-x_0$.

We initialize $B_{\beta^0}$ with $\theta_1$ by $B_{\beta^0}(k,x):=x+\frac{1}{N}m_{\theta_1}(k,x)$. First recalling that we are using the same $p_\textup{ref}$ (noise schedule) in SB-based models and SGMs, so the input $x_k$ of $B_{\beta^0}(k,\cdot)$ and $m_{\theta_1}(k,\cdot)$ follows the same distribution. Then, we rewrite the intermediate states at timestep $k$ and $k-1$ as 
\begin{subequations}
    \begin{align}
        x_k&=\left(1-\frac{k}{N}\right)x_0+\frac{k}{N}x_1, \label{eq:init_eq_1} \\ 
        x_{k-1}&=\left(1-\frac{k-1}{N}\right)x_0+\frac{k-1}{N}x_1, \label{eq:init_eq_2}
    \end{align}
\end{subequations}

\noindent and substitute the expression $x_0$ in Equation~(\ref{eq:init_eq_2}) by the corresponding term in Equation~(\ref{eq:init_eq_1}), leading to

\begin{equation}
    \begin{aligned}
        x_{k-1}&=x_k+\frac{1}{N-k}(x_k-x_1) \\
        &=x_k+\frac{1}{N-k}\left(1-\frac{k}{N}\right)(x_0-x_1) \\
        &=x_k+\frac{1}{N}(x_0-x_1),
    \end{aligned}
\end{equation}

\noindent where $x_{k-1}$ is the target of $B_{\beta^0}(k,x_k)$, and $(x_0-x_1)$ is exactly what $m_{\theta_1}$ predicts. For other noise schedules, \eg Variance-Preserving~\cite{ho2020denoising} and Variance-Exploding~\cite{song2019generative}, and reparameterization types, \eg noise ($\epsilon\sim p_\textup{prior}$) and data ($x_0\sim p_\textup{data}$), we can recalibrate similarly. Finally, we start from training $F_{\alpha^1}$.

We can also initialize both $B_{\beta^0}$ and $F_{\alpha^1}$ with two independent SGMs $m_{\theta_1}$ and $m_{\theta_2}$, which leads to faster convergence and better performance. We leave the details in the Appendix A.

\subsection{Necessity of Reparameterization}
\label{sec:lnitialization_original_dsb}

The training of our models greatly benefits from powerful initialization of pre-trained SGMs. This naturally leads to the question: can other SB-based methods also adopt and enjoy the same initialization strategy?

In IPMM, the training target in Equation~(\ref{eq:DSB_simplified_loss}) is the same as those used in common SGMs, which enables a drop-in replacement. However, in the case of DSB, the training target is Equation~(\ref{eq:DSB_original_loss}). The learning target is $\left(x_{k+1}+F_k^0(x_k)-F_k^0(x_{k+1})\right)$ for the backward network $B_{\beta^1}(k+1,x_{k+1})$ and $\left(x_{k}+B_{k+1}^1(x_{k+1})-B_{k+1}^1(x_{k})\right)$ for the forward network $F_{\alpha^1}(k,x_{k})$, which are significantly different from SGMs. Using a similar replacement like $B_{\beta^1}(k,x):=x+\frac{1}{N}m_{\theta_1}(k,x)$ will cause the \emph{out-of-domain} problem and accumulate error when calculating $B_{k+1}^1(x_{k})=B_{\beta^1}(k+1,x_k)=x_k+\frac{1}{N}m_{\theta_1}(k+1,x_k)$ in the training, as the input timestep $k+1$ is misaligned with the input state $x_k$.

Hence, the proposed initialization approaches are only enabled with reparameterization, and can not be adopted to other methods. 

\section{Experiment}

To thoroughly investigate the effectiveness of our proposed IPMM, IPTM, IPFM training objectives, and the initialization strategies, we conduct extensive experiments on various of tasks, including Gaussian fitting experiments, unconditional and conditional image generation, unpaired image-to-image translation.

\subsection{Experiment setup}

\noindent \textbf{Dataset and evaluation} For Gaussian experiments, we follow the common settings~\cite{de2021diffusion,shi2024diffusion}, where $p_\text{prior}=\mathcal{N}(-a,\mathbf{I})$, $p_\text{data}=\mathcal{N}(a,\mathbf{I})$, and $a\in\mathbb{R}^d$. The ground-truth $\pi^\textup{SB}$ can be analytically determined~\cite{de2021diffusion} and we evaluate $\text{KL}(\pi_t|\pi_t^\textup{SB})$ as a metric. For unconditional image generation, we adopt the CelebA~\cite{liu2015faceattributes} dataset. For class-conditional and text-conditional image generation, we use the ImageNet~\cite{russakovsky2015imagenet} and CUB-200~\cite{wah2011caltech} datasets, respectively. For unpaired image-to-image translation, we use the AFHQ~\cite{choi2020stargan}, Horse2Zebra~\cite{zhu2017unpaired} and Selfie2Anime~\cite{Kim2020U-GAT-IT:} datasets. We use Fréchet inception distance (FID)~\cite{heusel2017gans} as the metric for all experiments on images.

\noindent \textbf{Model and hyperparameter} For Gaussian experiments, we use a toy model with 10 fully connected layers. For unconditional image generation, we use ADM~\cite{dhariwal2021diffusion}. For conditional image generation and unpaired image-to-image translation, we borrow the network architecture of LDM~\cite{rombach2022high} and their corresponding VAEs. During training, we use Adam~\cite{kingma2014adam} optimizer with constant learning rate of $10^{-4}$, $\beta=\{0.9,0.99\}$, and no weight decay. We keep the batch size of 256 for all image sizes. We also follow~\cite{liu20232} to use the symmetric noise scheduling on $\gamma$.

\noindent \textbf{Baseline} We mainly compare with DSB~\cite{de2021diffusion}, IPML~\cite{vargas2021solving}, DSBM~\cite{shi2024diffusion}, BM$^2$~\cite{peluchetti2024bm}, and D-IMF~\cite{gushchin2024adversarial}. For image-to-image translation task, we additionally compare with I$^2$SB~\cite{liu20232}, BTTS~\cite{chen2023schrodinger}, BBDM~\cite{li2023bbdm}, and DDBM~\cite{zhou2023denoising}.

\begin{table*}[t]
\centering
\begin{tabular}{@{}c|cccccccc|cccc@{}}
\toprule
 & \multicolumn{8}{c|}{From scratch} & \multicolumn{4}{c}{With initialization} \\ \midrule
$d$ & DSB~\cite{de2021diffusion} & IPML~\cite{vargas2021solving} & DSBM~\cite{shi2024diffusion} & BM$^2$~\cite{peluchetti2024bm} & D-IMF~\cite{gushchin2024adversarial} & \underline{IPMM} & \underline{IPTM} & \underline{IPFM} & DSB~\cite{de2021diffusion} & \underline{IPMM} & \underline{IPTM} & \underline{IPFM} \\ \midrule
$2$ & 3.29 & 2.72 & 2.07 & 1.95 & 1.83 & 1.64 & 1.42 & \textbf{1.36} & 5.81 & 1.24 & 0.97 & \textbf{0.92} \\
$5$ & 7.76 & 6.91 & 3.86 & 4.03 & 3.44 & 2.85 & 2.64 & \textbf{2.55} & 11.17 & 1.53 & 1.18 & \textbf{1.16} \\
$20$ & 13.54 & 11.97 & 7.18 & 7.65 & 6.96 & 6.87 & \textbf{5.69} & 5.88 & 19.44 & 4.79 & 4.24 & \textbf{4.31} \\
$50$ & 32.91 & 28.42 & 12.37 & 11.98 & 11.03 & 10.34 & \textbf{9.67} & 9.93 & 45.26 & 8.56 & 8.23 & \textbf{7.97} \\ \bottomrule
\end{tabular}%
\vskip 0.1cm
\caption{Averaged $\textup{KL}(\pi_t|\pi_t^\textup{SB})$ at uniformly spaced $t$ of Gaussian experiments in different dimensions $d$. \underline{Underline} marks our models.}
\label{tab:gaussian_kl_comparison}
\vspace{0.2cm}
\end{table*}

\subsection{Gaussian experiments}

We first confirm that our method solves the SB problem with experiments on high-dimensional Gaussian distributions, following~\cite{de2021diffusion, shi2024diffusion}. In Table~\ref{tab:gaussian_kl_comparison}, our method trained from scratch predicts close to the analytically computed ground truth and outperforms other methods. We find our method needs less NFE and trains faster than DSB~\cite{de2021diffusion}, as discussed in Section~\ref{sec:simplified_target}.

We also study the gain of our initialization strategy. As shown in Table~\ref{tab:gaussian_kl_comparison}, using pretained SGMs as initialization consistently improves the performance. We also test DSB~\cite{de2021diffusion} and find that using SGMs as initialization could even adversely affect its performance due to the misaligned training target and timestep indexes, as discussed in Section~\ref{sec:lnitialization_original_dsb}.

Next, we compare our method, with and without initialization, to other SB-based models~\cite{de2021diffusion,vargas2021solving,shi2024diffusion,peluchetti2024bm,gushchin2024adversarial} on Gaussian experiments with different dimensions $d$. Table~\ref{tab:gaussian_kl_comparison} demonstrates that our method is better even without any initialization. We conjecture that modern reperameterization techniques could benefit and effectively stabilize the training process. Moreover, our method exhibits further improvements and achieves the best results with powerful initialization of pre-trained SGMs.

\begin{figure}[t]
  \begin{minipage}[c]{0.6\columnwidth}
    \centering
    \includegraphics[width=\linewidth]{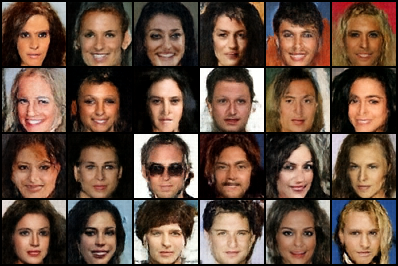}
    \captionof{figure}{Unconditional image generation results on CelebA.}
    \label{fig:results_celeba}
  \end{minipage}
  \hfill
  \begin{minipage}[c]{0.35\columnwidth}
    \centering
    \begin{tabular}{c|c}\hline
      Method & FID$\downarrow$ \\ \hline
      SGM & 23.29 \\
      DSB~\cite{de2021diffusion} & 31.78 \\
      DSBM~\cite{shi2024diffusion} & 20.45 \\
      \underline{IPFM} & \textbf{14.63} \\ \hline
      \end{tabular}
      \vskip 0.1cm
      \captionof{table}{FID results on CelebA. \underline{Underline} marks our models.}
      \label{tab:fid_celeba}
    \end{minipage}
    \vspace{0.5cm}
\end{figure}

\subsection{Unconditional image generation}

To validate whether our method can generalize to natural image data, we train models from scratch on the $64\times64$ CelebA~\cite{liu2015faceattributes} human face dataset with $50$ timesteps. In Figure~\ref{fig:results_celeba}, we illustrate the unconditional generation results of our IPFM model. It proves that our method is capable of generating images from Gaussian noises ($p_\textup{prior}=\mathcal{N}(0,1)$). We also report the commonly used FID~\cite{heusel2017gans} in Table~\ref{tab:fid_celeba} and compare to other methods.

\begin{table}[!t]
\centering
\begin{tabular}{@{}c|cccc@{}}
\toprule
Epochs & (Pre-trained SGM) & 2 & 4 & 6 \\ \midrule
\texttt{cat} $\rightarrow$ \texttt{dog} & 15.46 & 12.51 & 11.07 & \textbf{9.96} \\
\texttt{dog} $\rightarrow$ \texttt{cat} & 21.18 & 15.29 & 13.73 & \textbf{10.87} \\ \bottomrule
\end{tabular}%
\vskip 0.1cm
\caption{FID results on AFHQ dataset. Lower is better.}
\label{tab:afhq_fid}
\vspace{0.4cm}
\end{table}

\begin{figure}[!t]
    \centering
    \includegraphics[width=0.9\columnwidth]{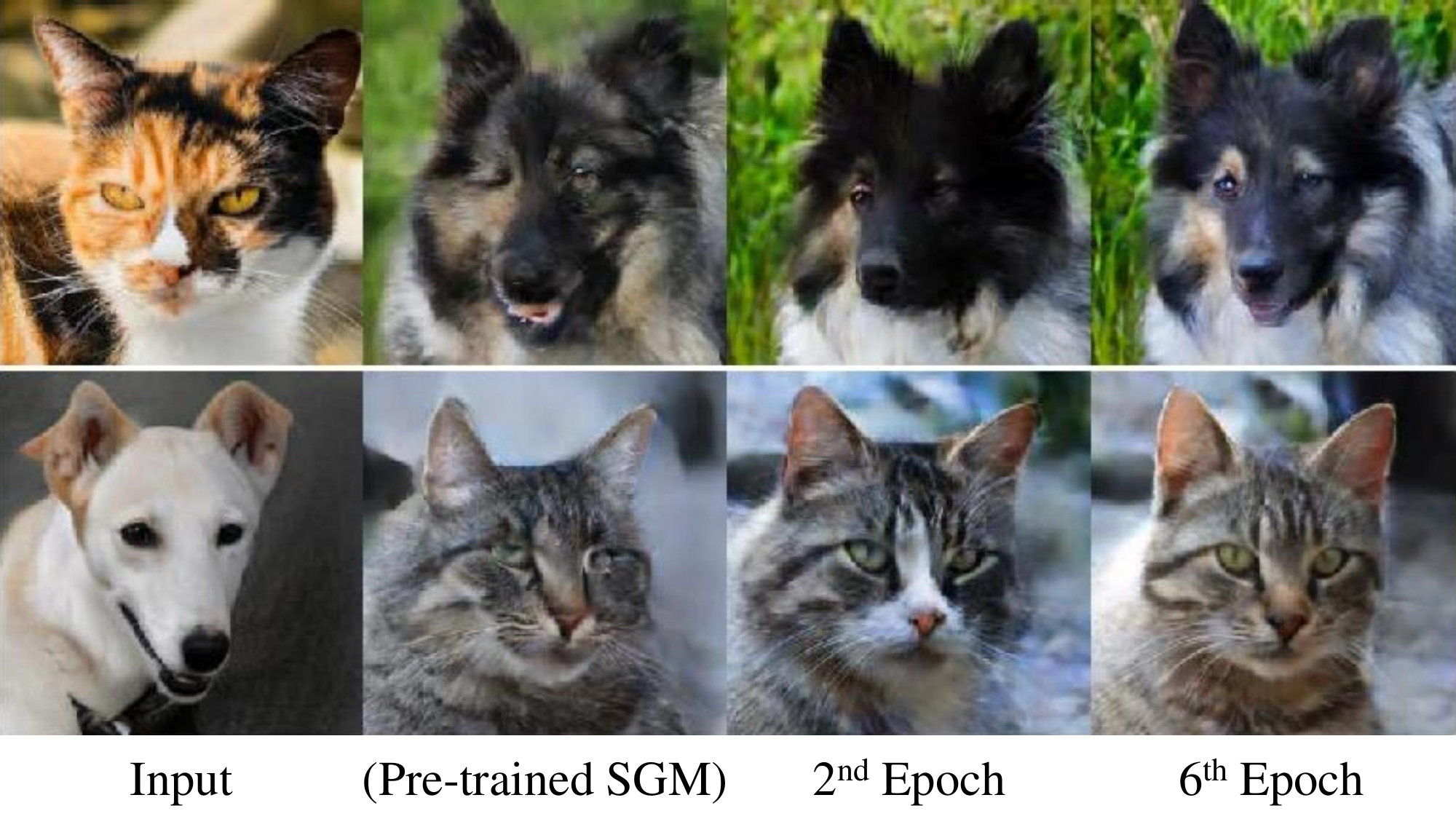}
    \vskip 0.1cm
    \caption{Generated images during training process on AFHQ.}
    \label{fig:afhq_epochs}
    \vspace{0.6cm}
\end{figure}

\begin{figure}[t]
    \centering
    \includegraphics[width=1.0\linewidth]{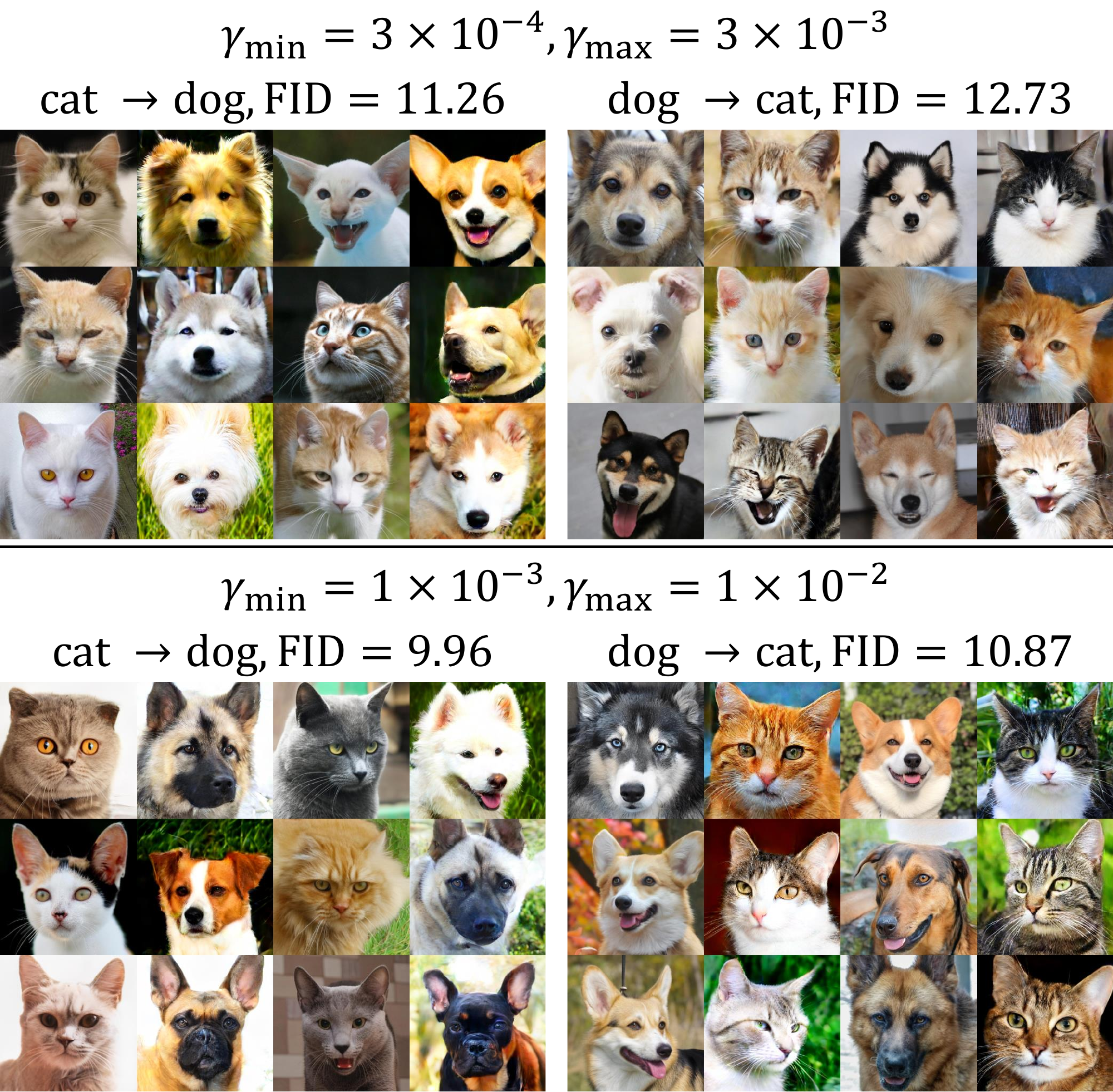}
    \vskip 0.1cm
    \caption{Translation samples on the unpaired $512\times512$ AFHQ dataset. For each pair, the left is source image and the right is generated image.}
    \label{fig:afhq-cases}
    \vspace{0.8cm}
\end{figure}

\begin{figure}[t]
    \centering
    \includegraphics[width=1.0\columnwidth]{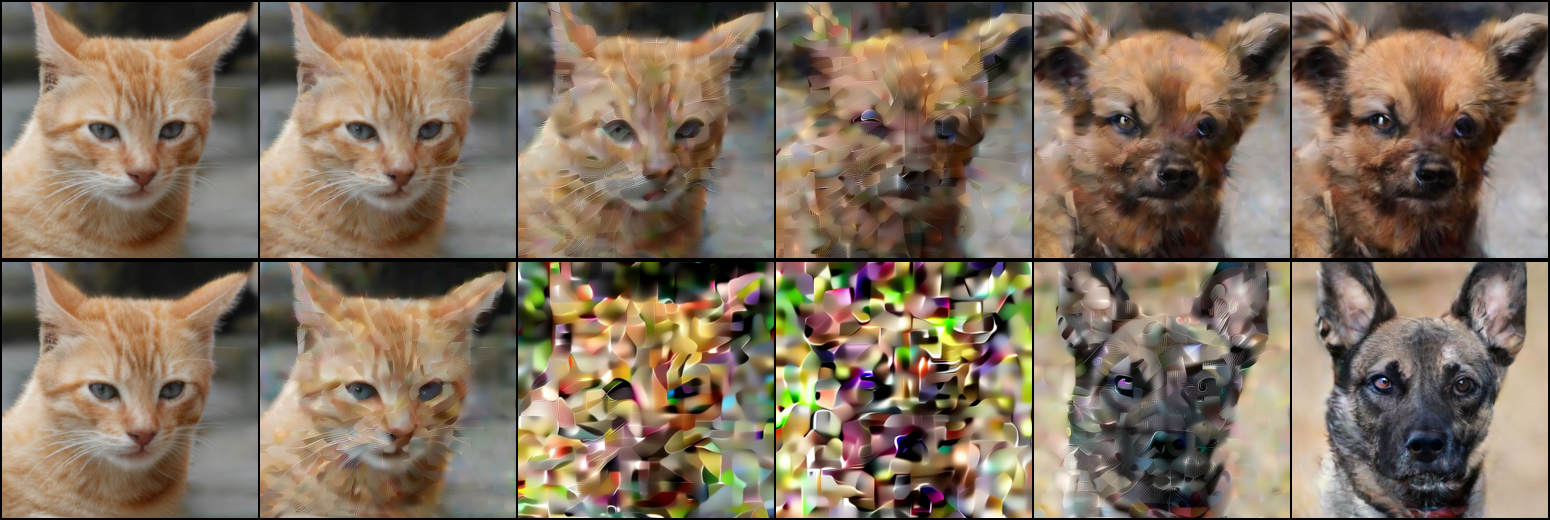}
    \vskip 0.1cm
    \caption{The translation trajectories of different $\gamma$ schedule.}
    \label{fig:gamma_visualize}
    \vspace{1.0cm}
\end{figure}

\begin{table*}[!t]
	\centering
	\resizebox{0.95\textwidth}{!}{%
		\hspace{-0.5cm}
		\begin{tabular}{@{}c|c|ccc|cccc|ccc@{}}
			\toprule
			& SGM & DSB~\cite{de2021diffusion} & IPML~\cite{vargas2021solving} & DSBM~\cite{shi2024diffusion} & I$^2$SB~\cite{liu20232} & BTTS~\cite{chen2023schrodinger} & BBDM~\cite{li2023bbdm} & DDBM~\cite{zhou2023denoising} & \underline{IPMM} & \underline{IPTM} & \underline{IPFM} \\ \midrule
			\texttt{cat} {\footnotesize$\rightarrow$} \texttt{dog} & 15.46 & 56.62 & 47.49 & 38.83 & 22.41 & 26.06 & 23.68 & 20.96 & 13.73 & 11.54 & \textbf{9.96} \\ 
			\texttt{dog} {\footnotesize$\rightarrow$} \texttt{cat} & 21.18 & 64.88 & 54.93 & 41.45 & 27.71 & 31.37 & 29.03 & 25.44 & 18.92 & 13.40 & \textbf{10.87} \\ \midrule
			\texttt{Horse} {\footnotesize$\rightarrow$} \texttt{Zebra} & 34.72 & 84.54 & 79.72 & 60.86 & 43.93 & 53.05 & 47.36 & 39.83 & 31.74 & 26.61 & \textbf{19.66} \\
			\texttt{Zebra} {\footnotesize$\rightarrow$} \texttt{Horse} & 35.43 & 89.81 & 77.11 & 62.06 & 46.27 & 51.93 & 46.50 & 42.11 & 32.31 & 27.95 & \textbf{18.83} \\  \midrule
			\texttt{Selfie} {\footnotesize$\rightarrow$} \texttt{Anime} & 31.37 & 71.01 & 65.49 & 50.46 & 39.04 & 43.77 & 43.16 & 36.55 & 28.86 & 21.01 & \textbf{16.65} \\ 
			\texttt{Anime} {\footnotesize$\rightarrow$} \texttt{Selfie} & 30.82 & 68.66 & 63.43 & 49.92 & 37.30 & 41.85 & 39.76 & 35.02 & 27.49 & 23.16 & \textbf{17.44} \\ \bottomrule
		\end{tabular}%
	}
	\vskip 0.1cm
	\caption{FID comparisons on AFHQ, Horse2Zebra, and Selfie2Anime datasets. \underline{Underline} marks our models.}
	\label{tab:fid_results}
    \vspace{0.2cm}
\end{table*}

\begin{table*}[t]
	\centering
		\begin{tabular}{@{}c|c|ccc|ccc@{}}
			\toprule
			\texttt{cat} {\footnotesize$\rightarrow$} \texttt{dog} & SGM & DSB~\cite{de2021diffusion} & IPML~\cite{vargas2021solving} & DSBM~\cite{shi2024diffusion} & \underline{IPMM} & \underline{IPTM} & \underline{IPFM} \\ \midrule
			Random Init. & 15.46 & 56.62 & 47.49 & 38.83 & 16.06 & 15.29 & \textbf{14.87} \\
			Pre-trained Init. & - & 54.57 & 48.30 & 41.28 & 13.73 & 11.54 & \textbf{9.96} \\ \bottomrule
		\end{tabular}%
        \\
	\vskip 0.1cm
		\begin{tabular}{@{}c|c|ccc|ccc@{}}
			\toprule
			\texttt{dog} {\footnotesize$\rightarrow$} \texttt{cat} & SGM & DSB~\cite{de2021diffusion} & IPML~\cite{vargas2021solving} & DSBM~\cite{shi2024diffusion} & \underline{IPMM} & \underline{IPTM} & \underline{IPFM} \\ \midrule
			Random Init. & 21.18 & 64.88 & 54.93 & 41.45 & 22.62 & 20.03 & \textbf{18.16} \\
			Pre-trained Init. & - & 60.20 & 51.77 & 39.00 & 18.92 & 13.40 & \textbf{10.87} \\ \bottomrule
		\end{tabular}%
	\vskip 0.1cm
	\caption{Ablation study of different initialization strategies. \underline{Underline} marks our models.}
	\label{tab:fid_init}
	\vspace{0.2cm}
\end{table*}

\subsection{Unpaired image-to-image translation}

We then scale up to the AFHQ~\cite{choi2020stargan}  $512\times512$ dataset to investigate unpaired translation between \texttt{cat} and \texttt{dog} subsets. We use a $\gamma$ schedule that starts at $\gamma_\textup{min}$, linearly grows to $\gamma_\textup{max}$ in the middle, and then linearly decays to $\gamma_\textup{min}$ by the end, following the the symmetric noise scheduling in~\cite{liu20232}.

To effectively train on the high-resolution data, we additionally propose to \emph{initialize $B_{\beta^1}$ and $F_{\alpha^1}$ separately}. We use two independent Bridge Models~\cite{li2023bbdm,zhou2023denoising} to initialize $B_{\beta^1}$ with $\theta_1$ that maps from \texttt{cat} to \texttt{dog}, and $F_{\alpha^1}$ with $\theta_2$ that maps from \texttt{dog} to \texttt{cat}. Specifically, we use $B_{\beta^1}(k,x):=x+\frac{1}{N}m_{\theta_1}(k,x)$ and $F_{\alpha^1}(k,x):=x+\frac{1}{N}m_{\theta_2}(N-k,x)$. We follow the first stage in~\cite{liu2022flow} and train two SGMs~\cite{lipman2022flow}, $m_{\theta_1}$ and $m_{\theta_2}$, until converged. Then, we train our IPFM models with initialization by the two pre-trained SGMs.

We report FID results in Table~\ref{tab:afhq_fid} and visualizations during the training process in Figure~\ref{fig:afhq_epochs}, showing that our method is capable of progressively improving image quality and surpassing the converged pre-trained SGMs that used as initialization. We provide more translation results in Figure~\ref{fig:afhq-cases}, which illustrates that our method is capable of producing rich details and realistic textures on high-resolution natural images. 

In addition, we find the $\gamma$ schedule controls the alignment of the appearance, \eg pose and color, between the source and generated pairs, and visualize the trajectories of different $\gamma$ in Figure~\ref{fig:gamma_visualize}. We also report results with different $\gamma$ schedules in Figure~\ref{fig:afhq-cases}, and conjecture that smaller $\gamma$ preserves more alignment at the cost of FID, while larger $\gamma$ produces better quality and diversity.

To thoroughly investigate the advantages of our method, we present comprehensive experiments in Table~\ref{tab:fid_results} on more datasets, including Horse2Zebra~\cite{zhu2017unpaired} and Selfie2Anime~\cite{Kim2020U-GAT-IT:} that are commonly used in unpaired image-to-image translations, and compare with more baselines, \eg I$^2$SB~\cite{liu20232}, BTTS~\cite{chen2023schrodinger}, BBDM~\cite{li2023bbdm}, and DDBM~\cite{zhou2023denoising}. The results indicate that our models not only show significant improvements over the pre-trained SGMs, but also consistently outperform other methods.

Finally, we conduct ablation studies on initialization settings in Table~\ref{tab:fid_init}. Our models enjoy notably improvements from the proposed initialization technique, while others are harmed possibly due to the misalignment issue discussed in Section~\ref{sec:lnitialization_original_dsb}.

\subsection{Class-to-image and text-to-image generation}

\begin{figure}[!t]
    \centering
    \includegraphics[width=1.0\columnwidth]{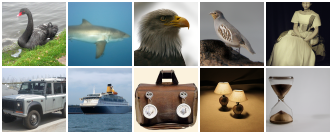}
    \vskip 0.1cm
    \caption{Generation results of class-to-image on ImageNet.}
    \label{fig:imagenet}
    \vspace{0.8cm}
\end{figure}

Conditional image generation is challenging yet imperative, as it holds greater practical significance in real-world applications. We further extend our method to class-to-image task on the ImageNet~\cite{russakovsky2015imagenet} 
$256\times256$ dataset and text-to-image task on the CUB-200~\cite{wah2011caltech} $256\times256$ dataset. The results presented in Table~\ref{tab:fid_conditional} and Figure~\ref{fig:imagenet} demonstrate that our models can also make notable improvements over the pre-trained SGMs, and generate vivid and diverse samples in the field of conditional image generation on large scale natural images. These promising results suggest that existing advanced diffusion models can also enjoy the improvements with our proposed methods.

\begin{table}[!t]
    \centering
    \resizebox{0.8\columnwidth}{!}{%
    \begin{tabular}{@{}c|cccc@{}}
    \toprule
    Epochs & (Pre-trained SGM) & 4 & 8 & 12 \\ \midrule
    ImageNet & 4.86 & 3.97 & 3.42 & \textbf{3.11} \\
    CUB-200 & 10.34 & 8.98 & 7.72 & \textbf{7.03} \\ \bottomrule
    \end{tabular}%
    }
    \vskip 0.1cm
    \caption{Fid results of class-to-image task on ImageNet and text-to-image task on CUB-200.}
    \label{tab:fid_conditional}
    \vspace{0.2cm}
\end{table}

\section{Conclusion}

This paper proposes reparameterization techniques, including IPMM, IPTM, and IPFM, which stabilize the training process and connect SB-based models to SGMs. This paper also proposes initialization strategies to leverage pre-trained SGMs, which incorporate powerful off-the-shelf models to accelerate convergence and further improve the performance of SGMs. Our work delves deep into the practical training of SB-based models, connects to and leverages advanced designs of concurrent SGMs, and bridges the research of the two fields for future developments of generative models.



\clearpage
\bibliography{mybibfile}




\section*{Supplementary material}

# Incorporating Pre-trained Diffusion Models in Solving the Schrödinger Bridge Problem

# Appendix

## Contents

## A  Initialization with SGMs

First, we explain how to use the three strategies proposed in Sec.V to initialize the network weights with pre-trained SGMs. For simplicity, consider we have two pre-trained SGMs that trained with Flow Matching [35] noise schedule, denoted as $\theta_1$ and $\theta_2$. $N$ denotes the number of total timesteps, and $k$ is index of an intermediate timestep. $\theta_1$ transits from $p_{\text{prior}}$ to $p_{\text{data}}$, *i.e.* the model $m_{\theta_1}$ with network parameter $\theta_1$ takes in $x_k = \left(1 - \frac{k}{N}\right)x_0 + \frac{k}{N}x_1$, where $x_0 \sim p_{\text{data}}$ and $x_1 \sim p_{\text{prior}}$, and learns $m_{\theta_1}(k, x_k) \to x_0 - x_1$. On the other hand, $\theta_2$ transits from $p_{\text{data}}$ to $p_{\text{prior}}$, *i.e.* the model $m_{\theta_2}$ with network parameter $\theta_2$ takes in $x_k = \frac{k}{N}x_0 + \left(1 - \frac{k}{N}\right)x_1$, where $x_0 \sim p_{\text{data}}$ and $x_1 \sim p_{\text{prior}}$, and learns $m_{\theta_2}(k, x_k) \to x_1 - x_0$.

**Init $B_{\beta^0}$.** We initialize $B_{\beta^0}$ with $\theta_1$, leaving $F_{\alpha^1}$ to be trained from scratch. Precisely, we use $B_{\beta^0}(k, x) := x + \frac{1}{N}m_{\theta_1}(k, x)$. First recalling that we are using the same $p_{\text{ref}}$ (noise schedule) in SB and SGMs, so the input $x_k$ of $B_{\beta^0}(k, \cdot)$ and $m_{\theta_1}(k, \cdot)$ follows the same distribution. Then, we rewrite the intermediate states at timestep $k$ and $k-1$ as

$$x_k = \left(1 - \frac{k}{N}\right)x_0 + \frac{k}{N}x_1, \tag{21a}$$

$$x_{k-1} = \left(1 - \frac{k-1}{N}\right)x_0 + \frac{k-1}{N}x_1, \tag{21b}$$

1

and substituting the expression $x_0$ in Equation (21b) by the corresponding term in Equation (21a) give

$$\begin{aligned} x_{k-1} &= x_k + \frac{1}{N-k}(x_k - x_1) \\ &= x_k + \frac{1}{N-k}\left(1 - \frac{k}{N}\right)(x_0 - x_1) \\ &= x_k + \frac{1}{N}(x_0 - x_1), \end{aligned} \quad (22)$$

where $x_{k-1}$ is the target of $B_{\beta^0}(k, x_k)$ and $(x_0 - x_1)$ is exactly what $m_{\theta_1}$ models. For other noise schedules, *e.g.* Variance-Preserving [26] and Variance-Exploding [51], and reparameterization types, *e.g.* noise ($\epsilon \sim p_{\text{prior}}$) and data ($x_0 \sim p_{\text{data}}$), we can also recalibrate similarly. Finally, we *i.e.* start from training $F_{\alpha^1}$.

***Init $B_{\beta^0}$ and $F_{\alpha^1}$ with same model***. we initialize both $B_{\beta^0}$ and $F_{\alpha^1}$ with the same model $\theta_1$. Precisely, we use $B_{\beta^0}(k, x) := x + \frac{1}{N} m_{\theta_1}(k, x)$ and $F_{\alpha^1}(k, x) := x - \frac{1}{N} m_{\theta_1}(k, x)$, which can be proved likewise. Despite the input distribution of $F_{\alpha^1}(k, x)$ might be slightly different from that of $m_{\theta_1}$, we find it still benefits and accelerates training, since $\pi^1$ is optimized to be close to $\pi^0 = p_{\text{ref}}$. Then, we can start training from the second epoch.

***Init $B_{\beta^0}$ and $F_{\alpha^1}$ separately***. we use two independent models to initialize $B_{\beta^0}$ with $\theta_1$, and $F_{\alpha^1}$ with $\theta_2$. Specifically, we use $B_{\beta^0}(k, x) := x + \frac{1}{N} m_{\theta_1}(k, x)$ and $F_{\alpha^1}(k, x) := x + \frac{1}{N} m_{\theta_2}(N - k, x)$ for initialization.

Finally, one may be confuse that SGMs assume the prior distribution $p_{\text{prior}}$ to be Gaussian, and how to train SGMs on two non-Gaussian distributions, such as the AFHQ `cat` and `dog` dataset? In fact, such SGMs is very similar to Bridge Models which have been thoroughly studied in [33, 59]. One slight different is that they need to train with paired data, while we train with unpaired data, *i.e.* randomly sampled pairs, as initialization. This also coincides with the initial stage of Rectified Flow [37]. Though the unpaired training is difficult to produce a good result, by iterating the finetune process, we can eventually get significantly better results.

## B Proof of IPMM

In Sec.IV, we propose an effective training target, *Iterative Propotional Mean-Matching (IPMM)*. Under mild assumptions, we can prove that the training loss of IPMM is identical to the original loss of DSB [13], while saving half of the NFEs, in the following proposition.

**Proposition 1.** *Assume that for any $n \in \mathbb{N}$ and $k \in \{0, 1, \ldots, N-1\}$, $\gamma_{k+1} > 0$, $q_{k|k+1}^n = \mathcal{N}\left(x_k; B_{k+1}^n(x_{k+1}), 2\gamma_{k+1}\mathbf{I}\right)$, $p_{k+1|k}^n = \mathcal{N}(x_{k+1}; F_k^n(x_k), 2\gamma_{k+1}\mathbf{I})$, where $B_{k+1}^n(x) = x + \gamma_{k+1} b_{k+1}^n(x)$ and $F_k^n(x) = x + \gamma_{k+1} f_k^n(x)$. Similar to Equation (4), we have*

$$\mathcal{L}'_{B_{k+1}^n} \approx \mathcal{L}_{B_{k+1}^n}, \mathcal{L}'_{F_k^{n+1}} \approx \mathcal{L}_{F_k^{n+1}}. \quad (23)$$

*Proof.* First recall that we use $f(x_k) \approx f(x_{k+1})$ in approximating the backward transition in Equation (4), we have $f_k^n(x_{k+1}) \approx f_k^n(x_k)$ and

$$\begin{aligned} &(x_{k+1} + F_k^n(x_k) - F_k^n(x_{k+1})) - x_k \\ &= (x_{k+1} + x_k + \gamma_{k+1} f_k^n(x_k) - (x_{k+1} + \gamma_{k+1} f_k^n(x_{k+1}))) - x_k \\ &= (x_k + \gamma_{k+1} f_k^n(x_k) - \gamma_{k+1} f_k^n(x_{k+1})) - x_k \\ &= \gamma_{k+1} f_k^n(x_k) - \gamma_{k+1} f_k^n(x_{k+1}) \\ &= f_k^n(x_k) - f_k^n(x_{k+1}) \\ &\approx 0, \end{aligned} \quad (24)$$

and thus $\mathcal{L}'_{B_{k+1}^n} \approx \mathcal{L}_{B_{k+1}^n}$. $\mathcal{L}'_{F_k^{n+1}} \approx \mathcal{L}_{F_k^{n+1}}$ can also be proved likewise. □

# C Proof of IPTM and IPFM

## C.1 Continuous-time DSB

We first introduce the continuous-time DSB. Given a reference measure $\mathbb{P} \in \mathscr{P}(\mathcal{C})$, the continuous-time SB problem aims to find

$$\Pi^* = \arg\min\{\mathrm{KL}(\Pi|\mathbb{P}) : \Pi \in \mathscr{P}(\mathcal{C}), \Pi_0 = p_{\mathrm{data}}, \Pi_T = p_{\mathrm{prior}}\}. \tag{25}$$

The continuous-time IPF iterations $(\Pi^n)_{n\in\mathbb{N}}$ where $\Pi^0 = \mathbb{P}$ for any $n \in \mathbb{N}$ can be defined in continuous time as

$$\begin{aligned}
\Pi^{2n+1} &= \arg\min\{\mathrm{KL}(\Pi|\Pi^{2n}) : \Pi \in \mathscr{P}(\mathcal{C}), \Pi_T = p_{\mathrm{prior}}\}, \\
\Pi^{2n+2} &= \arg\min\{\mathrm{KL}(\Pi|\Pi^{2n+1}) : \Pi \in \mathscr{P}(\mathcal{C}), \Pi_0 = p_{\mathrm{data}}\}.
\end{aligned} \tag{26}$$

DSB partitions the continuous time horizon $[0, T]$ into $N+1$ discrete timesteps $\{\bar{\gamma}_0, \bar{\gamma}_1, \ldots, \bar{\gamma}_N\}$ where $\bar{\gamma}_k = \sum_{n=1}^{k} \gamma_n$, $\bar{\gamma}_0 = 0$ and $\bar{\gamma}_N = T$. The forward process is $p^n_{\bar{\gamma}_{k+1}|\bar{\gamma}_k} = \mathcal{N}\left(x_{\bar{\gamma}_{k+1}}; x_{\bar{\gamma}_k} + \gamma_{k+1} f^n_{\bar{\gamma}_k}(x_{\bar{\gamma}_k}), 2\gamma_{k+1}\mathbf{I}\right)$ and the backward process is $q^n_{\bar{\gamma}_k|\bar{\gamma}_{k+1}} = \mathcal{N}\left(x_{\bar{\gamma}_k}; x_{\bar{\gamma}_{k+1}} + \gamma_{k+1} b^n_{\bar{\gamma}_{k+1}}(x_{\bar{\gamma}_{k+1}}), 2\gamma_{k+1}\mathbf{I}\right)$, where $f^n_{\bar{\gamma}_k}(x_{\bar{\gamma}_k})$ and $b^n_{\bar{\gamma}_{k+1}}(x_{\bar{\gamma}_{k+1}})$ are drift terms, and $p^n$ and $q^n$ are discretizations of $\Pi^{2n}$ and $\Pi^{2n+1}$, respectively. DSB uses two neural networks to approximate $B_{\beta^n}(\bar{\gamma}_k, x) \approx B^n_{\bar{\gamma}_k}(x) = x + \bar{\gamma}_k b^n_{\bar{\gamma}_k}(x)$ and $F_{\alpha^n}(\bar{\gamma}_k, x) \approx F^n_{\bar{\gamma}_k}(x) = x + \bar{\gamma}_{k+1} f^n_{\bar{\gamma}_k}(x)$.

## C.2 Schrödinger Bridge with SDE fomulation

Additionally, we introduce some notations in continuous-time SDEs. We rewrite the SDE formulation [32, 48] of SB as

$$\begin{aligned}
\mathrm{d}\mathbf{X}_t &= \left(f(\mathbf{X}_t, t) + g^2(t)\nabla\log\mathbf{\Psi}(\mathbf{X}_t, t)\right)\mathrm{d}t + g(t)\mathrm{d}\mathbf{W}_t, X_0 \sim p_{\mathrm{data}} \\
&:= h(\mathbf{X}_t, t)\mathrm{d}t + g(t)\mathrm{d}\mathbf{W}_t, X_0 \sim p_{\mathrm{data}},
\end{aligned} \tag{27a}$$

$$\begin{aligned}
\mathrm{d}\mathbf{X}_t &= \left(f(\mathbf{X}_t, t) - g^2(t)\nabla\log\hat{\mathbf{\Psi}}(\mathbf{X}_t, t)\right)\mathrm{d}t + g(t)\mathrm{d}\bar{\mathbf{W}}_t, X_0 \sim p_{\mathrm{data}} \\
&:= l(\mathbf{X}_t, t)\mathrm{d}t + g(t)\mathrm{d}\bar{\mathbf{W}}_t, X_1 \sim p_{\mathrm{prior}},
\end{aligned} \tag{27b}$$

where $h(\mathbf{X}_t, t)$ and $l(\mathbf{X}_t, t)$ are drift terms. Given the forward trajectory $x_{0:N} \sim p^n(x_{0:N})$ and the backward trajectory $x'_{0:N} \sim q^n(x_{0:N})$, $\{x_{\bar{\gamma}_i} := x_i\}_{i=0}^N$ and $\{x'_{\bar{\gamma}_i} := x'_i\}_{i=0}^N$ denote the trajectories of Equation (27a) and Equation (27b) at time $\{\bar{\gamma}_0, \bar{\gamma}_1, \ldots, \bar{\gamma}_N\}$, respectively. For simplicity, we denote the integration $\int_a^b h(\mathbf{X}_t, t)$ and $\int_a^b l(\mathbf{X}_t, t)$ as $\tilde{h}(a, b)$ and $\tilde{l}(a, b)$, respectively.

## C.3 Posterior estimation

Similar to DDPM[26], we derive an estimation of the posterior transitions for $p^n_{k|k+1,0}(x_k|x_{k+1}, x_0)$ and $q^n_{k+1|k,N}(x_{k+1}|x_k, x_N)$. For simplicity, we denote $\bar{\gamma}_k = \sum_{n=1}^{k}\gamma_n$ and $\bar{\gamma}_0 = 0$. We have the following proposition.

**Proposition 2.** *Assume $\bar{\gamma}_N = 1$. Given the forward and backward trajectories $x_{0:N} \sim p^n(x_{0:N}) = p^n_0(x_0)\prod_{k=0}^{N-1} p^n_{k+1|k}(x_{k+1}|x_k)$ and $x'_{0:N} \sim q^n(x_{0:N}) = q^n_N(x_N)\prod_{k=0}^{N-1} p^n_{k|k+1}(x_k|x_{k+1})$. Assume $f_t(\mathbf{X}_t) = -\alpha\mathbf{X}_t$, $g(t) = \sqrt{2}$,*

$$\frac{1}{\bar{\gamma}_k}\tilde{h}(0, \bar{\gamma}_k) \approx \frac{1}{\bar{\gamma}_{k+1}}\tilde{h}(0, \bar{\gamma}_{k+1}), \quad \frac{1}{1 - \bar{\gamma}_k}\tilde{l}(\bar{\gamma}_k, 1) \approx \frac{1}{1 - \bar{\gamma}_{k+1}}\tilde{l}(\bar{\gamma}_{k+1}, 1). \tag{28}$$

*We have*

$$q^n_{k|k+1}(x_k|x_{k+1}) \approx p^n_{k|k+1,0}(x_k|x_{k+1}, x_0) = \mathcal{N}(x_k; \mu^n_{k+1}(x_{k+1}, x_0), \sigma_{k+1}\mathbf{I}), \tag{29a}$$

$$p^{n+1}_{k+1|k}(x_{k+1}|x_k) \approx q^n_{k+1|k,N}(x'_{k+1}|x'_k, x'_N) = \mathcal{N}(x'_{k+1}; \tilde{\mu}^n_k(x'_k, x'_N), \tilde{\sigma}_{k+1}\mathbf{I}), \tag{29b}$$

3

$$\mu_{k+1}^n(x_{k+1}, x_0) \approx x_{k+1} + \frac{\gamma_{k+1}}{\bar{\gamma}_{k+1}}(x_0 - x_{k+1}), \sigma_{k+1} = \frac{2\gamma_{k+1}\bar{\gamma}_k}{\bar{\gamma}_{k+1}}, \tag{30a}$$

$$\tilde{\mu}_k^n(x'_k, x'_N) \approx x'_k + \frac{\gamma_{k+1}}{1-\bar{\gamma}_k}(x'_N - x'_k), \tilde{\sigma}_{k+1} = \frac{2\gamma_{k+1}(1-\bar{\gamma}_{k+1})}{1-\bar{\gamma}_k}. \tag{30b}$$

It is worth noticing that we use $f_t(\mathbf{X}_t) = -\alpha \mathbf{X}_t$ and $g(t) = \sqrt{2}$ in consistent with the original DSB [13]. We prove Proposition 2 with the following Lemma 1, Lemma 2, and Lemma 3.

**Lemma 1.** *Given the forward and backward trajectories $x_{0:N} \sim p^n(x_{0:N}) = p_0^n(x_0)\prod_{k=0}^{N-1} p_{k+1|k}^n(x_{k+1}|x_k)$ and $x'_{0:N} \sim q^n(x_{0:N}) = q_N^n(x_N)\prod_{k=0}^{N-1} p_{k|k+1}^n(x_k|x_{k+1})$. Suppose the IPF in Equation (26) achieve the optimal state. Then we have*

$$q_{k|k+1}^n(x_k|x_{k+1}) \approx p_{k|k+1,0}^n(x_k|x_{k+1}, x_0), \tag{31a}$$

$$p_{k+1|k}^{n+1}(x_{k+1}|x_k) \approx q_{k+1|k,N}^n(x'_{k+1}|x'_k, x'_N). \tag{31b}$$

*Proof.* We first prove Equation (31a). In the $(2n+1)^{\text{th}}$ iteration, IPF aims to solve

$$\begin{aligned}
\text{KL}(q^n|p^n) &= \mathbb{E}_{q^n}\left(\log \frac{q^n(x_N)\prod_{k=0}^{N-1} q^n(x_k|x_{k+1})}{p^n(x_0)p^n(x_N|x_0)\prod_{k=1}^{N-1} p^n(x_k|x_{k+1}, x_0)}\right) \\
&= \mathbb{E}_{q^n}\left(\underbrace{\log \frac{q^n(x_0|x_1)}{p^n(x_0)}}_{\text{reconstruction}} + \underbrace{\log \frac{q^n(x_N)}{p^n(x_N|x_0)}}_{\text{prior matching}} + \sum_{k=1}^{N-1} \underbrace{\log \frac{q^n(x_k|x_{k+1})}{p^n(x_k|x_{k+1}, x_0)}}_{\text{denoising matching}}\right).
\end{aligned} \tag{32}$$

On the minimization of Equation (32), we have $q_{k|k+1}^n(x_k|x_{k+1}) \approx p_{k|k+1,0}^n(x_k|x_{k+1}, x_0)$. We can prove Equation (31b) likewise. □

**Lemma 2.** *Assume $\bar{\gamma}_N = 1$. Given the forward trajectory $x_{0:N} \sim p^n(x_{0:N}) = p_0^n(x_0)\prod_{k=0}^{N-1} p_{k+1|k}^n(x_{k+1}|x_k)$ and the and backward trajectory $x'_{0:N} \sim q^n(x_{0:N}) = q_N^n(x_N)\prod_{k=0}^{N-1} p_{k|k+1}^n(x_k|x_{k+1})$. Assume $f_t(\mathbf{X}_t) = -\alpha \mathbf{X}_t$ and $g(t) = \sqrt{2}$. We have*

$$p_{k|k+1,0}^n(x_k|x_{k+1}, x_0) = \mathcal{N}(x_k; \mu_{k+1}^n(x_{k+1}, x_0), \sigma_{k+1}\mathbf{I}), \tag{33a}$$

$$q_{k+1|k,N}^n(x'_{k+1}|x'_k, x'_N) = \mathcal{N}(x'_{k+1}; \tilde{\mu}_k^n(x'_k, x'_N), \tilde{\sigma}_{k+1}\mathbf{I}), \tag{33b}$$

*where*

$$\mu_{k+1}^n(x_{k+1}, x_0) = x_{k+1} + \frac{\gamma_{k+1}}{\bar{\gamma}_{k+1}}(x_0 - x_{k+1}) - \tilde{h}(\bar{\gamma}_k, \bar{\gamma}_{k+1}) + \frac{\gamma_{k+1}}{\bar{\gamma}_{k+1}}\tilde{h}(0, \bar{\gamma}_{k+1}),$$

$$\sigma_{k+1} = \frac{2\gamma_{k+1}\bar{\gamma}_k}{\bar{\gamma}_{k+1}}, \tag{34a}$$

$$\tilde{\mu}_k^n(x'_k, x'_N) = x'_k + \frac{\gamma_{k+1}}{1-\bar{\gamma}_k}(x'_N - x'_k) + \tilde{l}(\bar{\gamma}_k, \bar{\gamma}_{k+1}) - \frac{\gamma_{k+1}}{1-\bar{\gamma}_k}\tilde{l}(\bar{\gamma}_k, 1),$$

$$\tilde{\sigma}_{k+1} = \frac{2\gamma_{k+1}(1-\bar{\gamma}_{k+1})}{1-\bar{\gamma}_k}. \tag{34b}$$

*Proof.* We first prove Equation (33a) and Equation (34a). Denote for simplicity

$$\begin{aligned}
\tilde{h}_1 &= \tilde{h}(\bar{\gamma}_k, \bar{\gamma}_{k+1}) = \int_{\bar{\gamma}_k}^{\bar{\gamma}_{k+1}} \left(f(x_t, t) + g^2(t)\nabla \log \mathbf{\Psi}(x_t, t)\right) dt, \\
\tilde{h}_2 &= \tilde{h}(\bar{\gamma}_0, \bar{\gamma}_k) = \int_{\bar{\gamma}_0}^{\bar{\gamma}_k} \left(f(x_t, t) + g^2(t)\nabla \log \mathbf{\Psi}(x_t, t)\right) dt, \\
\tilde{h}_3 &= \tilde{h}(\bar{\gamma}_0, \bar{\gamma}_{k+1}) = \int_{\bar{\gamma}_0}^{\bar{\gamma}_{k+1}} \left(f(x_t, t) + g^2(t)\nabla \log \mathbf{\Psi}(x_t, t)\right) dt,
\end{aligned} \tag{35}$$

4

and we have
$$p_{k+1|k}^n(x_{k+1}|x_k) = \mathcal{N}\left(x_{k+1}; x_k + \tilde{h}_1, 2\gamma_{k+1}\mathbf{I}\right),$$
$$p_{k|0}^n(x_k|x_0) = \mathcal{N}\left(x_k; x_0 + \tilde{h}_2, 2\bar{\gamma}_k\mathbf{I}\right), \tag{36}$$
$$p_{k+1|0}^n(x_{k+1}|x_0) = \mathcal{N}\left(x_{k+1}; x_0 + \tilde{h}_3, 2\bar{\gamma}_{k+1}\mathbf{I}\right),$$

$$\begin{aligned}
&p_{k|k+1,0}^n(x_k|x_{k+1}, x_0)\\
&= \frac{p_{k+1|k}^n(x_{k+1}|x_k)p_{k|0}^n(x_k|x_0)}{p_{k+1|0}^n(x_{k+1}|x_0)}\\
&= \frac{\mathcal{N}\left(x_{k+1}; x_k + \tilde{h}_1, 2\gamma_{k+1}\mathbf{I}\right)\mathcal{N}\left(x_k; x_0 + \tilde{h}_2, 2\bar{\gamma}_k\mathbf{I}\right)}{\mathcal{N}\left(x_{k+1}; x_0 + \tilde{h}_3, 2\bar{\gamma}_{k+1}\mathbf{I}\right)}\\
&\propto \exp\left\{-\frac{1}{2}\left[\frac{\left(x_{k+1} - x_k - \tilde{h}_1\right)^2}{2\gamma_{k+1}} + \frac{\left(x_k - x_0 - \tilde{h}_2\right)^2}{2\bar{\gamma}_k} - \frac{\left(x_{k+1} - x_0 - \tilde{h}_3\right)^2}{2\bar{\gamma}_{k+1}}\right]\right\}\\
&= \exp\left\{-\frac{1}{4}\left[\left(\frac{1}{\gamma_{k+1}} + \frac{1}{\bar{\gamma}_k}\right)x_k^2 - 2\left(\frac{x_{k+1} - \tilde{h}_1}{\gamma_{k+1}} + \frac{x_0 + \tilde{h}_2}{\bar{\gamma}_k}\right)x_k + C_1\right]\right\}\\
&= \exp\left\{-\frac{1}{2}\frac{\left[x_k - \left(\frac{\bar{\gamma}_k}{\bar{\gamma}_{k+1}}(x_{k+1} - \tilde{h}_1) + \frac{\gamma_{k+1}}{\bar{\gamma}_{k+1}}(x_0 + \tilde{h}_2)\right)\right]^2}{\frac{2\gamma_{k+1}\bar{\gamma}_k}{\bar{\gamma}_{k+1}}} + C_2\right\}\\
&= \exp\left\{-\frac{1}{2}\frac{\left[x_k - \left(x_{k+1} + \frac{\gamma_{k+1}}{\bar{\gamma}_{k+1}}(x_0 - x_{k+1}) - \tilde{h}_1 + \frac{\gamma_{k+1}}{\bar{\gamma}_{k+1}}\tilde{h}_3\right)\right]^2}{\frac{2\gamma_{k+1}\bar{\gamma}_k}{\bar{\gamma}_{k+1}}} + C_2\right\}\\
&\propto \mathcal{N}(x_k; x_{k+1} + \frac{\gamma_{k+1}}{\bar{\gamma}_{k+1}}(x_0 - x_{k+1}) - \tilde{h}_1 + \frac{\gamma_{k+1}}{\bar{\gamma}_{k+1}}\tilde{h}_3, \frac{2\gamma_{k+1}\bar{\gamma}_k}{\bar{\gamma}_{k+1}}\mathbf{I}),
\end{aligned} \tag{37}$$

where $C_1$ and $C_2$ are constants up to $\{x_0, x_{k+1}, \tilde{h}_1, \tilde{h}_2, \tilde{h}_3\}$. We can also prove Equation (33b) and Equation (34b) likewise. □

**Lemma 3.** *Assume that*
$$\frac{1}{\bar{\gamma}_k}\tilde{h}(0, \bar{\gamma}_k) \approx \frac{1}{\bar{\gamma}_{k+1}}\tilde{h}(0, \bar{\gamma}_{k+1}), \quad \frac{1}{1 - \bar{\gamma}_k}\tilde{l}(\bar{\gamma}_k, 1) \approx \frac{1}{1 - \bar{\gamma}_{k+1}}\tilde{l}(\bar{\gamma}_{k+1}, 1). \tag{38}$$

*Then, we have*
$$\tilde{h}(\bar{\gamma}_k, \bar{\gamma}_{k+1}) - \frac{\gamma_{k+1}}{\bar{\gamma}_{k+1}}\tilde{h}(0, \bar{\gamma}_{k+1}) \approx 0, \tag{39a}$$
$$\tilde{l}(\bar{\gamma}_k, \bar{\gamma}_{k+1}) - \frac{\gamma_{k+1}}{1 - \bar{\gamma}_k}\tilde{l}(\bar{\gamma}_k, 1) \approx 0. \tag{39b}$$

*Proof.* We first prove Equation (39a). We have
$$\begin{aligned}
&\tilde{h}(\bar{\gamma}_k, \bar{\gamma}_{k+1}) - \frac{\gamma_{k+1}}{\bar{\gamma}_{k+1}}\tilde{h}(0, \bar{\gamma}_{k+1})\\
&= \bar{\gamma}_k\left(\frac{1}{\bar{\gamma}_k}\tilde{h}(\bar{\gamma}_k, \bar{\gamma}_{k+1}) - \left(\frac{1}{\bar{\gamma}_k} - \frac{1}{\bar{\gamma}_{k+1}}\right)\tilde{h}(0, \bar{\gamma}_{k+1})\right)\\
&= \bar{\gamma}_k\left(-\frac{1}{\bar{\gamma}_k}\tilde{h}(0, \bar{\gamma}_k) + \frac{1}{\bar{\gamma}_{k+1}}\tilde{h}(0, \bar{\gamma}_{k+1})\right)\\
&\approx 0.
\end{aligned} \tag{40}$$

We can also prove Equation (39b) likewise. □

Lemma 1, Lemma 2, and Lemma 3 together conclude the proof of Proposition 2.

### C.4 Reparameterization

According to the forward and backward propagation of DSB, we have

$$\begin{aligned}
x_{\bar{\gamma}_{k+1}} &= F^n_{\bar{\gamma}_k}(x_{\bar{\gamma}_k}) + \sqrt{2\gamma_{k+1}}\epsilon = x_{\bar{\gamma}_k} + \gamma_{k+1}f^n_{\bar{\gamma}_k}(x_{\bar{\gamma}_k}) + \sqrt{2\gamma_{k+1}}\epsilon, \\
x_{\bar{\gamma}_k} &= B^n_{\bar{\gamma}_{k+1}}(x_{\bar{\gamma}_{k+1}}) + \sqrt{2\gamma_{k+1}}\epsilon = x_{\bar{\gamma}_{k+1}} + \gamma_{k+1}b^n_{\bar{\gamma}_{k+1}}(x_{\bar{\gamma}_{k+1}}) + \sqrt{2\gamma_{k+1}}\epsilon.
\end{aligned} \quad (41)$$

Comparing Equation (41) with Equation (27), we have

$$\begin{aligned}
x_{\bar{\gamma}_{k+1}} &= x_{\bar{\gamma}_k} + \int_{\bar{\gamma}_k}^{\bar{\gamma}_{k+1}} \left[\left(f(x_t, t) + g^2(t)\nabla \log \Psi(x_t, t)\right) \mathrm{d}t + g(t)\mathrm{d}\mathbf{W}_t\right] \\
&= x_{\bar{\gamma}_k} + \int_{\bar{\gamma}_k}^{\bar{\gamma}_{k+1}} \left(f(x_t, t) + g^2(t)\nabla \log \Psi(x_t, t)\right) \mathrm{d}t + \sqrt{2\gamma_{k+1}}\epsilon, \\
x_{\bar{\gamma}_k} &= x_{\bar{\gamma}_{k+1}} + \int_{\bar{\gamma}_{k+1}}^{\bar{\gamma}_k} \left[\left(f(x_t, t) - g^2(t)\nabla \log \hat{\Psi}(x_t, t)\right) \mathrm{d}t + g(t)\mathrm{d}\bar{\mathbf{W}}_t\right] \\
&= x_{\bar{\gamma}_{k+1}} + \int_{\bar{\gamma}_{k+1}}^{\bar{\gamma}_k} \left(f(x_t, t) - g^2(t)\nabla \log \hat{\Psi}(x_t, t)\right) \mathrm{d}t + \sqrt{2\gamma_{k+1}}\epsilon,
\end{aligned} \quad (42)$$

where $g(t) \equiv \sqrt{2}$ in [13] and $\epsilon \sim \mathcal{N}(0, 1)$. By comparing Equation (41) and Equation (42), we can find that DSB essentially learns

$$\begin{aligned}
f^n_{\bar{\gamma}_k}(x_{\bar{\gamma}_k}) &\approx \frac{1}{\gamma_{k+1}} \int_{\bar{\gamma}_k}^{\bar{\gamma}_{k+1}} \left(f(x_t, t) + g^2(t)\nabla \log \Psi(x_t, t)\right) \mathrm{d}t, \\
b^n_{\bar{\gamma}_{k+1}}(x_{\bar{\gamma}_{k+1}}) &\approx \frac{1}{\gamma_{k+1}} \int_{\bar{\gamma}_{k+1}}^{\bar{\gamma}_k} \left(f(x_t, t) - g^2(t)\nabla \log \hat{\Psi}(x_t, t)\right) \mathrm{d}t,
\end{aligned} \quad (43)$$

where Equation (27) can be viewed as the limit at $\gamma_{k+1} \to 0$. However, $\mathrm{d}\mathbf{X}_t$ can be noisy to learn as $\mathbf{X}_t$ is sampled from SDE, which involves iteratively adding Gaussian noise, and it is hard for the network to converge in our practical settings. Given Proposition 2 and the observations in Equation (43), we propose IPTM and IPFM in Sec.IV, inspired by prevalent SGMs. Here we provide some intuitive understandings.

**Iterative Propotional Terminus-Matching (IPTM)** We use two neural networks to learn $\tilde{F}_{\alpha^n}(\bar{\gamma}_k, x_{\bar{\gamma}_k}) \approx \tilde{F}^n_{\bar{\gamma}_k}(x_{\bar{\gamma}_k})$ and $\tilde{B}_{\beta^n}(\bar{\gamma}_{k+1}, x_{\bar{\gamma}_{k+1}}) \approx \tilde{B}^n_{\bar{\gamma}_{k+1}}(x_{\bar{\gamma}_{k+1}})$ s.t.

$$\begin{aligned}
\tilde{B}^n_{\bar{\gamma}_{k+1}}(x_{\bar{\gamma}_{k+1}}) &\approx x_0 = x_{\bar{\gamma}_{k+1}} + \int_{\bar{\gamma}_{k+1}}^{\bar{\gamma}_0} \left(f(x_t, t) - g^2(t)\nabla \log \hat{\Psi}(x_t, t)\right) \mathrm{d}t, \\
\tilde{F}^n_{\bar{\gamma}_k}(x_{\bar{\gamma}_k}) &\approx x_1 = x_{\bar{\gamma}_k} + \int_{\bar{\gamma}_k}^{\bar{\gamma}_N} \left(f(x_t, t) + g^2(t)\nabla \log \Psi(x_t, t)\right) \mathrm{d}t,
\end{aligned} \quad (44)$$

and the training loss is

$$\begin{aligned}
\mathcal{L}_{\tilde{B}^n_{k+1}} &= \mathbb{E}_{p^n_{0,k+1}} \left[\left\|\tilde{B}^n_{k+1}(x_{k+1}) - x_0\right\|^2\right], \\
\mathcal{L}_{\tilde{F}^{n+1}_k} &= \mathbb{E}_{q^n_{k,N}} \left[\left\|\tilde{F}^{n+1}_k(x_k) - x_N\right\|^2\right].
\end{aligned} \quad (45)$$

Intuitively, the networks learn to predict terminal points $x_0 \sim p_{\text{data}}$ and $x_1 \sim p_{\text{prior}}$ given $x_t$. This recovers DDPM [26] when $\nabla \log \Psi(x_t, t) = 0$, which means we are adding randomly drawn Gaussian noise to sampled data.

**Iterative Proportional Flow-Matching (IPFM)** We use two neural networks to learn $\tilde{f}_{\alpha^n}(\bar{\gamma}_k, x_{\bar{\gamma}_k}) \approx \tilde{f}^n_{\bar{\gamma}_k}(x_{\bar{\gamma}_k})$ and $\tilde{b}_{\beta^n}(\bar{\gamma}_{k+1}, x_{\bar{\gamma}_{k+1}}) \approx \tilde{b}^n_{\bar{\gamma}_{k+1}}(x_{\bar{\gamma}_{k+1}})$ s.t.

$$\tilde{b}^n_{\bar{\gamma}_{k+1}}(x_{\bar{\gamma}_{k+1}}) \approx \frac{1}{\bar{\gamma}_{k+1} - \bar{\gamma}_0} \int_{\bar{\gamma}_{k+1}}^{\bar{\gamma}_0} \left( f(x_t, t) - g^2(t) \nabla \log \hat{\Psi}(x_t, t) \right) dt,$$

$$\tilde{f}^n_{\bar{\gamma}_k}(x_{\bar{\gamma}_k}) \approx \frac{1}{\bar{\gamma}_N - \bar{\gamma}_k} \int_{\bar{\gamma}_k}^{\bar{\gamma}_N} \left( f(x_t, t) + g^2(t) \nabla \log \Psi(x_t, t) \right) dt, \quad (46)$$

and the training loss is

$$\mathcal{L}_{\tilde{b}^n_{k+1}} = \mathbb{E}_{p^n_{0,k+1}} \left[ \left\| \tilde{b}^n_{k+1}(x_{k+1}) - \frac{x_0 - x_{k+1}}{\bar{\gamma}_{k+1}} \right\|^2 \right],$$

$$\mathcal{L}_{\tilde{f}^{n+1}_k} = \mathbb{E}_{q^n_{k,N}} \left[ \left\| \tilde{f}^{n+1}_k(x_k) - \frac{x_N - x_k}{1 - \bar{\gamma}_k} \right\|^2 \right]. \quad (47)$$

Intuitively, the networks learn the vector field that points from $x_t$ to $x_0$ and $x_1$. This recovers Flow Matching [35] when $\nabla \log \Psi(x_t, t) = 0$, which means we are adding randomly drawn Gaussian noise to sampled data.

In summary, IPTM and IPFM learns the integration of Equation (27) over the interval. Comparing to the original fomulation of DSB, learning the integration diminishes the noise along the trajetories, and leads to more efficient, stable and accurate training.

## D Related works

### D.1 Score-based Generative Models

Score-based Generative Models (SGMs) are first introduced in [50], and SMLD [51, 52] later establishes a theoretical framework using Stochastic Differential Equations. Independently, DDPM [26] provides an explanation in terms of Markov chains and introduces the noise prediction reparameterization. Subsequent works [16] achieve state-of-the-art results and outperform GANs [20] in image generation task. Successive improvements to SGMs are made through network design [27, 43], training strategy [3, 21, 29, 45], noise schedule [34, 54], sampler [29, 38], guidance method [16, 25], and reparameterization [47]. Extensions of SGMs to discrete spaces are also explored [2, 6, 22, 53]. SGMs are widely applied in image generations [16, 45], image superresolution [46], 3D generation [28], and video generation [5, 40].

### D.2 Flow and Bridge Models

Neural ODE [8] first introduces continuous-time interpretation of Normalizing Flows [30, 41] into the generation community. Flow Matching (FM) [35] proposes a simulation-free version of Neural ODE with inspirations from SGMs. Rectified Flow [37] further proposes a reflow strategy and applies it to general distribution transformations beyond Gaussian. CFM [55] investigates the role of couplings and use minibatch Optimal Transport (OT) pairs in training. RFM [7] further expands Flow Matching to accommodate general Riemannian manifolds. These methods employ Ordinary Differential Equations (ODEs) to learn the transport map between two distributions.

Other works introduce stochasticity and leverages SDE to model the generation process. [1] is the pioneer work and InDI [14] uses Bridge Matching and explores its applications on the task of image translation and restoration. BBDM [33] and DDBM [59] provide theoretical studies, leveraging Doob's h-transforms [17] to build the stochastic bridge between two arbitrary distributions. These works reveal the relations between Bridge Models and SGMs.

Though some of these methods possess the potentials to transform between non-Gaussian distributions, some limitations still exist. Some works [59] can only be trained on paired datasets and others [37] may require an iterative modeling process.

### D.3 Generative modeling based on Schrödinger Bridge

Recently, new methods emerge and solve the Schrödinger Bridge problem by discretizing the state-space [10, 44], approximating potential functions using regression [4, 15], performing kernel density estimation [42]. DSB [13] decomposes classical IPF algorithm and leverages several designs akin to SGMs. The work IPML [56] also simplifies IPF, and proves that solving the Schrödinger Bridge is equivalent to an autoregressive maximum likelihood estimation objective. SB-FBSDE [9] proposes to maximize likelihood using forward-backward SDEs theory and develops an approach to simultaneously optimize two networks. More recent work ENOT [23] develops an end-to-end learning algorithm based on the saddle point reformulation.

Another kind of works try to solve Schrödinger Bridge by approximating the boundary distribution as a mixture of Dirac delta distributions. $I^2$SB [36] first adopts it into image-to-image translations. Bridge-TTS [11] studies the tractable Schrödinger Bridge between Gaussian-smoothed paired data and extends the scheme to different noise schedules and ODE samplers. However, these two methods can only be applied to paired datasets and did not generalize to unpaired tasks.

### D.4 Generative modeling based on Optimal Transport and Entropic Optimal Transport

Some researchers treat generative modeling from a more general perspective of distribution transportation and Optimal Transport (OT). LSOT [19, 49] solves a classic dual Entropic Optimal Transport (EOT) problem [18] with two neural networks and learns a more network for the barycentric projection. SCONES [12] combines SGMs and the potentials from LSOT to sample from Langevin dynamics with a score model. NOT [31] reformulates the weak OT dual into a max-min optimization and adopts two networks to learn the transport map and potentials. Energy-guided EOT [39] employs energy-based modeling to solve weak EOT dual and learns only one network. However, these methods have not been tested on larger scales or high-resolution images.

## E Compatibility with non-uniform timestep designs

Recent works [24, 57, 58] have shown that different timesteps contribute differently to the quality of generation of the diffusion model, and certain non-uniform designs on the timesteps are proposed to improve the fidelity of samples. As a vivid example for bridging diffusion and SB models and importing the advanced designs of diffusion models to SB models, we conduct an additional experiment on the compatibility with non-uniform timestep designs in Table 1. As shown, our method is orthogonal to and also benefits from these designs.

| `cat → dog` | IPFM | +min-SNR [24] | +B-TTDM [58] | +SpeeD [57] |
|---|---|---|---|---|
| Random Init. | 14.87 | 13.94 | 13.46 | 13.68 |
| Pre-trained Init. | 9.96 | 9.43 | 9.07 | 9.20 |

| `dog → cat` | IPFM | +min-SNR [24] | +B-TTDM [58] | +SpeeD [57] |
|---|---|---|---|---|
| Random Init. | 18.16 | 17.25 | 16.35 | 16.21 |
| Pre-trained Init. | 10.87 | 9.89 | 9.36 | 9.28 |

Table 1: Experiments with non-uniform timestep designs.

## F Additional results

We present close-up results on the AFHQ dataset in Figure 1 and Figure 2. It demonstrates that our method generates realistic appearances and rich details on high-dimensional space.

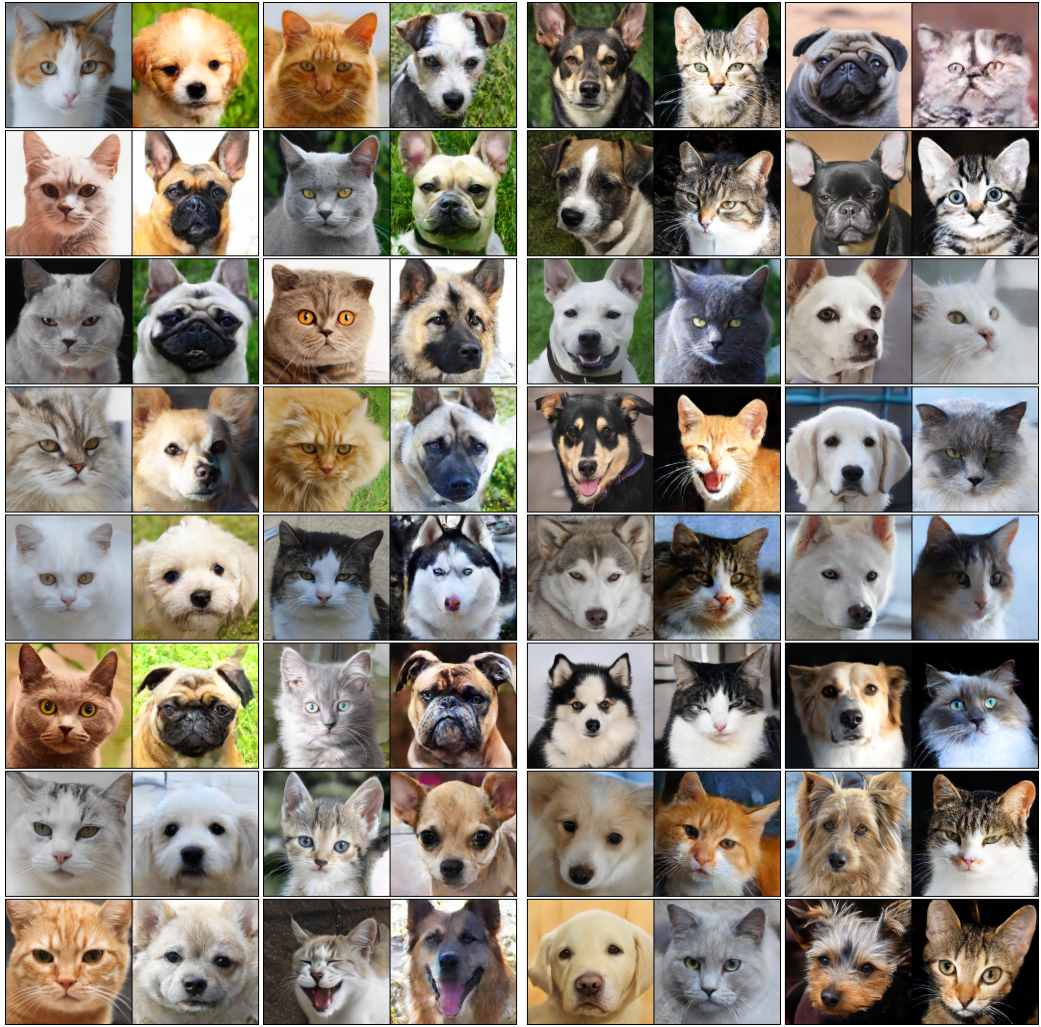

Figure 1: More results on unpaired image-to-image translation between dog and cat. Left: cat to dog; right: dog to cat. For each case, the left is the input image and the right is the generated result from our model.

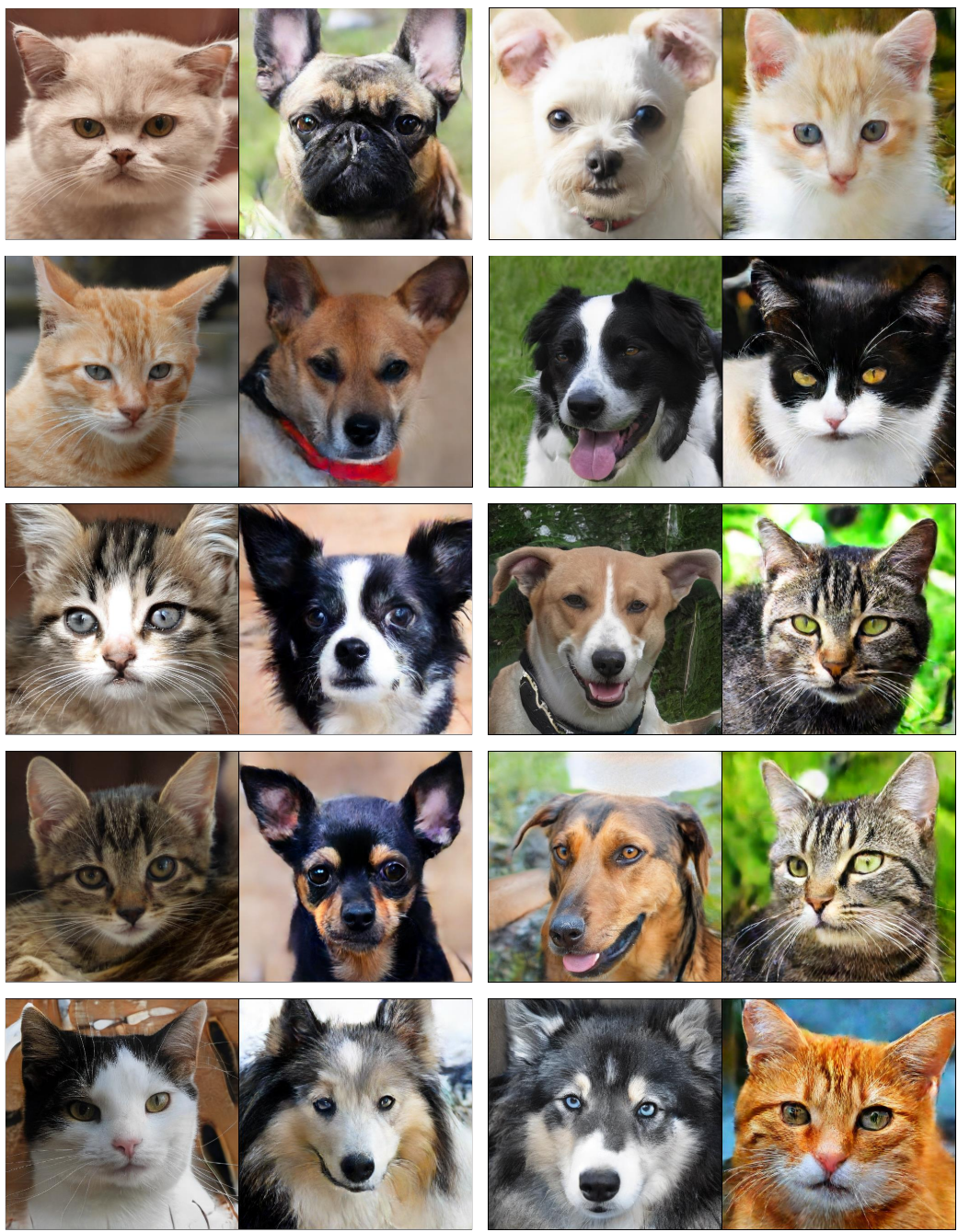

Figure 2: More results on unpaired image-to-image translation between dog and cat. Left: cat to dog; right: dog to cat. For each case, the left is the input image and the right is the generated result from our model.



\end{document}